\documentclass{article}

\usepackage[preprint]{neurips_2020}
\usepackage[utf8]{inputenc} 
\usepackage[T1]{fontenc}    
\usepackage{hyperref}       
\usepackage{url}            
\usepackage{booktabs}      
\usepackage{amsfonts}       
\usepackage{nicefrac}       
\usepackage{microtype}      
\usepackage{tikz}
\usepackage{pgfplots}
\usetikzlibrary{arrows}
\usepackage{algorithm}
\usepackage{algorithmic}
\usepackage{amsmath}
\usepackage{microtype}
\usepackage{multicol}
\usepackage{times}
\usepackage{helvet}
\usepackage{courier}
\usepackage{url}
\usepackage{graphicx}
\usepackage{natbib}
\usepackage{caption}
\usepackage{bm}
\usepackage{mathtools}
\usepackage{xfrac}
\usepackage{tikz}
\usepackage{amsmath,amsfonts,bm,amssymb,amsthm}
\usepackage[colorinlistoftodos,textsize=scriptsize]{todonotes}
\usepackage{soul}
\newcommand{\transpose}{\mathsf{T}}

\newcommand{\ver}{\mathrm{VER}}

\newcommand{\threshold}{t}

\newcommand{\identity}{y}

\renewcommand{\vec}{\bm}

\DeclarePairedDelimiterX{\infdivx}[2]{(}{)}{%
  #1\;\delimsize\|\;#2
}

\newcommand{\sampleElement}{x}
\newcommand{\labelElement}{y}

\newcommand{\seed}{q}

\newcommand{\bigOh}{O}

\newcommand{\sampleSpace}{\mathcal{X}}

\newcommand{\labelSpace}{\mathcal{Y}}
\newcommand{\sampleDist}{\mathcal{D}}

\newcommand{\classifier}{\mathrm{f}}
\newcommand{\predictor}{\classifier}
\newcommand{\generativeModel}{\mathrm{h}}

\newcommand{\normalDist}{\mathcal{N}}

\newcommand{\placeholderDist}{\nu}

\newcommand{\promptSpace}{\mathcal{P}}
\newcommand{\prompt}{p}

\newcommand{\metric}{\rho}

\usepackage{dsfont}

\newcommand{\indicatorFunction}{\mathds{1}}

\DeclarePairedDelimiter{\abs}{\lvert}{\rvert}

\newcommand{\sample}{\vec{\sampleElement}}

\newcommand{\realNumbers}{\mathbb{R}}

\newcommand{\dimension}{d}
\newcommand{\embeddingDimension}{k}
\newcommand{\trainingData}{S}

\newcommand{\placeholder}{z}

\newcommand{\definedAs}{\triangleq}

\newcommand{\groupA}{\mathsf{a}}
\newcommand{\groupB}{\mathsf{b}}

\newcommand{\group}{\mathsf{g}}
\newcommand{\groups}{\mathsf{G}}

\newcommand\MTkillspecial[1]{
\bgroup
\catcode`\&=9
\let\\\relax%
\scantokens{#1}%
\egroup
}
\DeclarePairedDelimiter\brparen
\lparen\rparen
\reDeclarePairedDelimiterInnerWrapper\brparen{star}{
\mathopen{#1\vphantom{\MTkillspecial{#2}}\kern-\nulldelimiterspace\right.}
#2
\mathclose{\left.\kern-\nulldelimiterspace\vphantom{\MTkillspecial{#2}}#3}}

\newcommand{\embeddingFunc}{\mathrm{f}}

\usepackage{clipboard}

\newcommand{\totalVariation}{\mathrm{TV}}

\makeatletter
\newcommand\RedeclareMathOperator{%
  \@ifstar{\def\rmo@s{m}\rmo@redeclare}{\def\rmo@s{o}\rmo@redeclare}%
}
\newcommand\rmo@redeclare[2]{%
  \begingroup \escapechar\m@ne\xdef\@gtempa{{\string#1}}\endgroup
  \expandafter\@ifundefined\@gtempa
     {\@latex@error{\noexpand#1undefined}\@ehc}%
     \relax
  \expandafter\rmo@declmathop\rmo@s{#1}{#2}}
\newcommand\rmo@declmathop[3]{%
  \DeclareRobustCommand{#2}{\qopname\newmcodes@#1{#3}}%
}
\@onlypreamble\RedeclareMathOperator
\makeatother

\RedeclareMathOperator{\Pr}{{\mathbb{P}}}

\usepgfplotslibrary{groupplots}
\usepackage{cleveref}
\theoremstyle{definition}
\newtheorem{definition}{Definition}
\theoremstyle{plain}
\newtheorem{theorem}{Theorem}

\newtheorem{proposition}{Proposition}
\theoremstyle{definition}
\newtheorem{nullhypothesis}{Null Hypothesis}
\crefname{nullhypothesis}{null hypothesis}{null hypotheses}
\Crefname{nullhypothesis}{Null hypothesis}{Null hypotheses}
\usepackage{booktabs}
\usepackage{subcaption}
\usepackage{multirow}
\usepackage{makecell}
\newcommand*\samethanks[1][\value{footnote}]{\footnotemark[#1]}

\title{Limitations of Face Image Generation}

\author{\hspace*{-1cm}
Harrison Rosenberg\thanks{equal contribution}, Shimaa Ahmed\samethanks, Guruprasad V Ramesh\samethanks, Ramya Korlakai Vinayak, Kassem Fawaz\\
  Electrical and Computer Engineering Department\\
  University of Wisconsin--Madison \\
  \hspace*{-1cm}\url{hrosenberg@ece.wisc.edu}, \url{{ahmed27,viswanathanr,kfawaz}@wisc.edu}, \url{ramya@ece.wisc.edu}
  }

\begin{document}
\maketitle
\begin{abstract}
 Text-to-image diffusion models have achieved widespread popularity due to their unprecedented image generation capability. In particular, their ability to synthesize and modify human faces has spurred research into using generated face images in both training data augmentation and model performance assessments. In this paper, we study the efficacy and shortcomings of generative models in the context of face generation. Utilizing a combination of qualitative and quantitative measures, including embedding-based metrics and user studies, we present a framework to audit the characteristics of generated faces conditioned on a set of social attributes. We applied our framework on faces generated through state-of-the-art text-to-image diffusion models. We identify several limitations of face image generation that include faithfulness to the text prompt, demographic disparities, and distributional shifts. Furthermore, we present an analytical model that provides insights into how training data selection contributes to the performance of generative models. Our survey data and analytics code can be found online\footnote{\url{https://github.com/wi-pi/Limitations_of_Face_Generation}}.

\end{abstract}

\section{Introduction}

\begin{figure}[t]
    \centering 
        \includegraphics[width=\columnwidth]{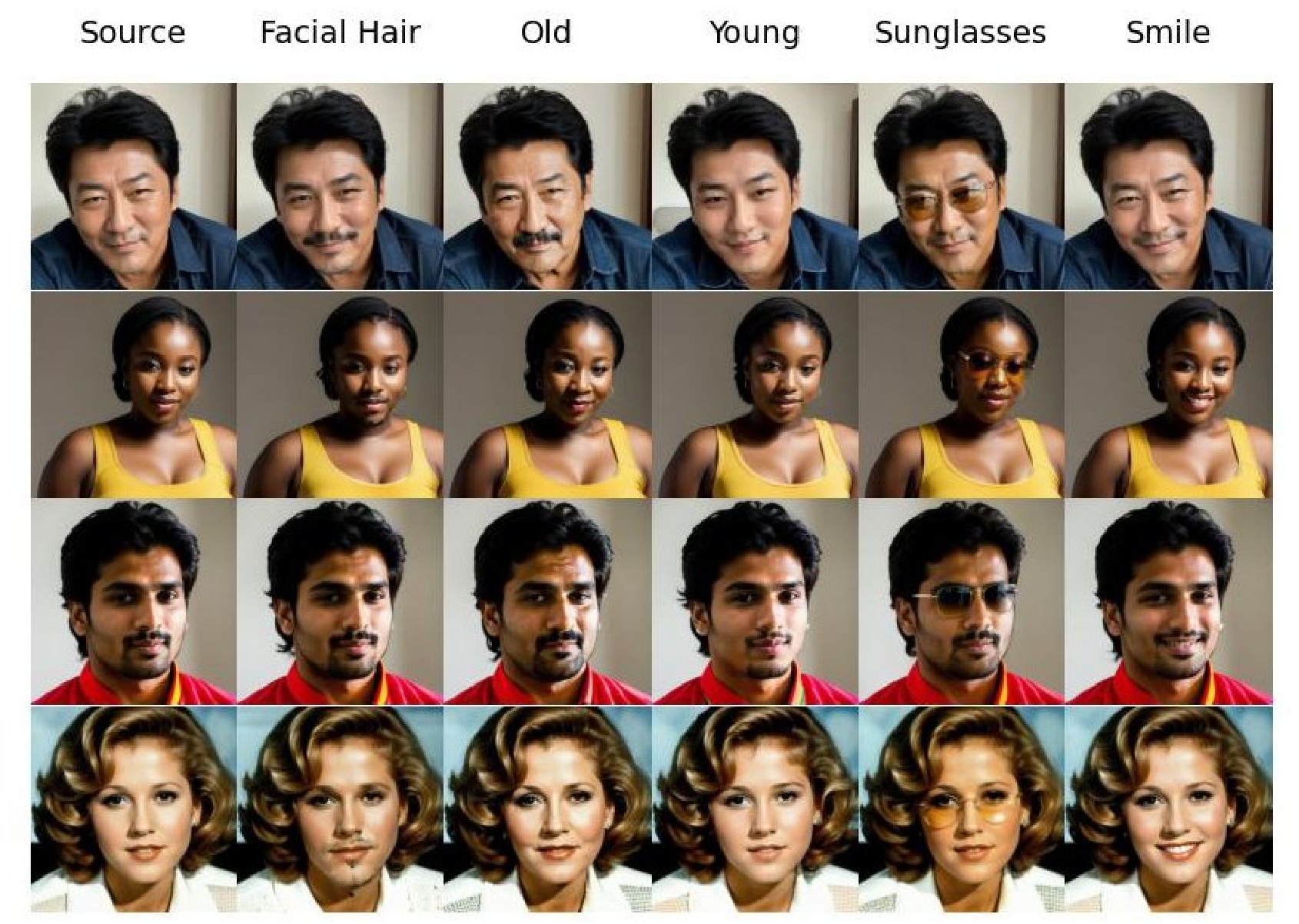}
    \caption{Samples of the non-celebrities dataset using Realism model for four demographic groups: `East Asian Male', `Black Female', `Indian Male', and `White Female'. `Source' refers to images generated from Realism using the prompt template. The second to sixth columns show transformed images when an attribute is applied to `Source' using SEGA.}
    \label{fig:non-celeb-data}
\end{figure}

Text-to-image~(TTI) diffusion models have become popular due to their unprecedented image-generation capability. Taking a textual prompt as input, these models generate realistic images which align with user intentions. Their ability to synthesize and modify human faces has spurred research into using generated face images in training data augmentation and model performance assessments~\cite{dixit2017aga, trabucco2023effective}. For example, face recognition systems can benefit from synthetic datasets that exhibit more demographic diversity than existing natural datasets~\cite{smith2023balancing, friedrich2023fair}. 

This work analyzes the quality of synthetic datasets for facial recognition applications and whether they exhibit demographic disparities. Achieving this objective requires generating a large set of identities belonging to diverse demographic groups and generating multiple (different) faces for each identity. Existing diffusion models are incapable of meeting this objective for two reasons. First, aligning the generated faces with the provided prompt is challenging~\cite{tsaban2023ledits, brack2023sega}. Second, generating multiple faces with the same identity in a one-shot fashion is typically infeasible~\cite{tsaban2023ledits, brack2023sega}. Limited research exists in this space. Previous works either optimize the diffusion model to a particular demographic group, generate faces without a notion of identity, or limit their objectives to frequency analysis of the demographics of the generated images~\cite{perera2023analyzing}. 

In this work, we propose a new framework to generate synthetic face images, as shown in~\cref{fig:non-celeb-data}.
Our face-generation pipeline takes as input demographic attributes, applies custom prompts to generate identities for each demographic attribute, and utilizes image editing models~\cite{brack2023sega} to generate diversified faces for each identity. The resulting dataset, which we manually verify, resembles a natural face image dataset, albeit demographically balanced by design.

We then apply a three-pronged approach to assess the synthetic face image dataset's quality through face verification~\cite{schroff2015facenet}, quantitative quality metrics~\cite{ruiz2023dreambooth}, and a user study. Our evaluation shows that generated images exhibit demographic disparities in the eyes of face recognition systems. Results from our user studies show disparity in quality of the generated faces for different demographics, with images belonging to majority demographics rated as higher quality. We also study the efficacy of edit correctness metrics built on CLIP and DINO~\cite{brooks2022instructpix2pix,zhang2023MagicBrush}. We find these metrics do not correlate with human preferences in facial semantic changes. Research is needed to develop perceptually aligned metrics.

Finally, our findings suggest that methods intended to mitigate bias exhibit demographic disparities in the quality of generated images. Through an analytical model, we show that generative models mimic the demographic disparities in the existing data, typically sampled from the Internet. We also develop sample complexity and data sampling conditions to overcome this inherent bias.

To the best of our knowledge, this paper is the first work that: (1) \emph{provides an end-to-end pipeline}, utilizing a TTI diffusion model, to generate batches of synthetic faces annotated with fine-grained attributes; (2) \emph{evaluates the quality of large-scale batch-generated faces} using a user study; and (3) \emph{assesses the fidelity} of recently proposed TTI quality metrics on face images.

\section{Related Work}
In the following section, we describe recent works in the context of synthetic face image generation and associated biases.

\subsection{Synthetic Face Image Generation}
TTI diffusion models, such as DALL-E~\cite{ramesh2021zero} and Stable Diffusion~\cite{rombach2021highresolution}, rely on internal randomness to generate high-quality examples through denoising steps. They employ CLIP~\cite{radford2021learning} or its variants as text encoders. Thus, 
a text prompt is sufficient to control the output of a TTI diffusion model.

Two challenges arise in prompt-based face generation. The first is aligning the generated faces with the provided prompt~\cite{tsaban2023ledits, brack2023sega}. The second is generating multiple faces belonging to the same identity in a one-shot fashion~\cite{tsaban2023ledits, brack2023sega}. There exist methods to better control image generation. These methods include segmentation masks and inpainting~\cite{zhang2023MagicBrush}; text-inversion, which learns a text token that corresponds to certain image concept~\cite{gal2022image}; model fine-tuning and embedding optimization~\cite{kawar2023imagic}. While these techniques are generally effective, they are unsuitable for large-scale generation of diverse faces. 
They often disrupt the fast and natural interface that differentiates TTI diffusion models.


In this work, we aim to analyze the synthetic faces generated by TTI diffusion models. This objective requires generating a large set of identities belonging to diverse demographic groups and generating multiple (different) faces for each identity.  We devise a novel pipeline that employs semantic guided attention (SEGA)~\cite{brack2023sega}, fixed seeds, and specialized prompts. The pipeline, described within the Framework section, depends on neither inversion nor fine-tuning.

\subsection{Bias in Face Image Generation}

Recent works~\cite{friedrich2023fair, seshadri2023bias, smith2023balancing} have studied the bias of TTI face generation by analyzing the proportions of demographics in generated images. Seshadri et al. found that generative models amplify the discrepancies in training data~\cite{seshadri2023bias}. One example is gender-occupation bias, where Stable Diffusion can generate highly biased face distributions from a gender-neutral prompt about occupations. Friedrich et al. mitigated these biases with a post-processing technique called Fair Diffusion~\cite{friedrich2023fair}. When the user inputs their prompt, a model detects the potential bias in the prompt and steers the output to a fairer region, leveraging a lookup table of instructions and the semantic image-editing technique SEGA~\cite{brack2023sega}. Similarly, Smith et al. utilized InstructPix2Pix~\cite{brooks2022instructpix2pix}, an instruction-based image editing model, to edit existing images to be demographically balanced. While this dataset debiasing technique results in finer-grained control over demographic attributes, it introduces a distribution shift between natural and synthetic images. It also stacks the biases of different models~\cite{smith2023balancing}. 

Luccioni et al. performed a different bias characterization that relies on correlating model outputs in the embedding space with social attributes~\cite{luccioni2023stable}. The authors found three popular TTI models are biased toward masculine and white concepts. Struppek et al. studied another source of bias resulting from non-Latin scripts~\cite{struppek2022biased}. They found that using special non-Latin characters better exposes the internal biases of models and proposed using homoglyphs to mitigate this bias. Mu{\~n}oz et al. analyzed the bias in relatively older face generation models trained on the CelebA and FFHQ datasets~\cite{munoz2023uncovering}. Using quantitative metrics, including demographic frequencies, face recognition verification, and Fréchet inception distance, they found that the generative models are biased.

These conclusions are consistent with earlier GAN literature, where Maluleke et al. found them to generate racially biased distributions of faces~\cite{maluleke2022studying}. Maluleke et al. went one step further by analyzing the quality of generated faces through a user study, where generated faces from minority groups (e.g., Black) exhibited lower quality.

In summary, existing works focus primarily on frequency analysis to characterize bias in TTI models, propose embedding-based metrics to evaluate the quality of generated images, and utilize synthetic data to mitigate bias. In our work, we characterize the synthetic datasets, showing that methods intended to mitigate bias exhibit demographic disparities in the quality of generated images. We go beyond frequency analysis by rating image quality in a user study. We also utilize the user study results to assess embedding-based metrics in characterizing the quality of the generated images.

\section{Framework}
We develop a framework, as depicted in \cref{fig:pipeline}, to audit the characteristics of generated face images. This framework consists of choosing the demographic conditions, prompting diffusion models to generate identities according to these conditions, followed by evaluating the generated images both quantitatively and qualitatively. 

\subsection{Notation}
We first prescribe the notation used within this paper. There exists a sample space $\sampleSpace \subseteq \realNumbers^{\dimension}$ and label set $\labelSpace$. A sample $\sample \in \sampleSpace$ is a $\dimension$-dimensional vector.  If $\sample$ is an RGB image, then $\dimension$ equals $3 \times h \times w$, which corresponds to the number of channels multiplied by the number of pixels in the image. In the context of face recognition, we assume each face image $\sample$ depicts an identity $\identity \in \labelSpace$. A face recognition model $\predictor: \sampleSpace \to \labelSpace$ is trained on a finite dataset $\trainingData \in \sampleSpace \times \labelSpace$. $\trainingData$ is drawn i.i.d. from distribution $\sampleDist$. Sometimes, when clear from context, $\trainingData$ refers to an unlabeled dataset. A metric embedding network $\embeddingFunc_\embeddingDimension: \sampleSpace \to \realNumbers^\embeddingDimension$ is often internal to deep-network based classifiers. Metric embedding network $\embeddingFunc_\embeddingDimension$ maps inputs to a $\embeddingDimension$-dimensional embedding space. If two samples have low pairwise distance, they are assumed to be more similar in the associated label space.

We analyze disparities in generative models across social attributes. To analyze these disparities, we examine synthetic face image quality and the performance of generated images in face recognition tasks. A common class of social attributes is demographics. With respect to demographics, we use terminology consistent with Buolamwini and Gebru~\cite{buolamwini2018gender}, a work among the most cited in the space of face recognition fairness. Face images are annotated by sex and ethnicity. Sex annotations are ``Male'' and ``Female.'' Ethnicity annotations are ``White,'' ``Black,'' ``East Asian,'' and ``Indian.'' The set of demographic groups is denoted as $\groups$, where $\group$ is a placeholder for a demographic group in $\groups$. In this paper, we study eight demographic groups, corresponding to sex-ethnicity combinations. 

Given a text prompt $\prompt$ from the space of prompts $\promptSpace$, a text-to-image model $\generativeModel_\seed: \promptSpace \to \sampleSpace$ returns the image prescribed by its textual prompt $\prompt$, where the random seed $\seed$ is a real number. Because diffusion models have internal randomness, each $\seed$ generates a different realization of the same prompt $\prompt$. In our framework, we encode the identity $\identity$ and its demographic group $\group$ in the textual prompt $\prompt$, and we vary the seed $\seed$ to generate multiple images of the same identity. We use a fixed set of seeds to ensure the reproducibility of generated images.

\subsection{Generative Models}
We generate synthetic faces using two TTI Diffusion models: the open-source Stable Diffusion v2.1 model by Stability AI and the finetuned Realistic Vision Model\footnote{\url{https://huggingface.co/SG161222/Realistic_Vision_V4.0_noVAE}}, hereafter referred to as \emph{SDv2.1} and \emph{Realism}, respectively. We analyze the images generated by these models individually to assess their efficacy in face-generation pipelines.

\begin{figure}[t]
    \centering 
    \includegraphics[width=\columnwidth]{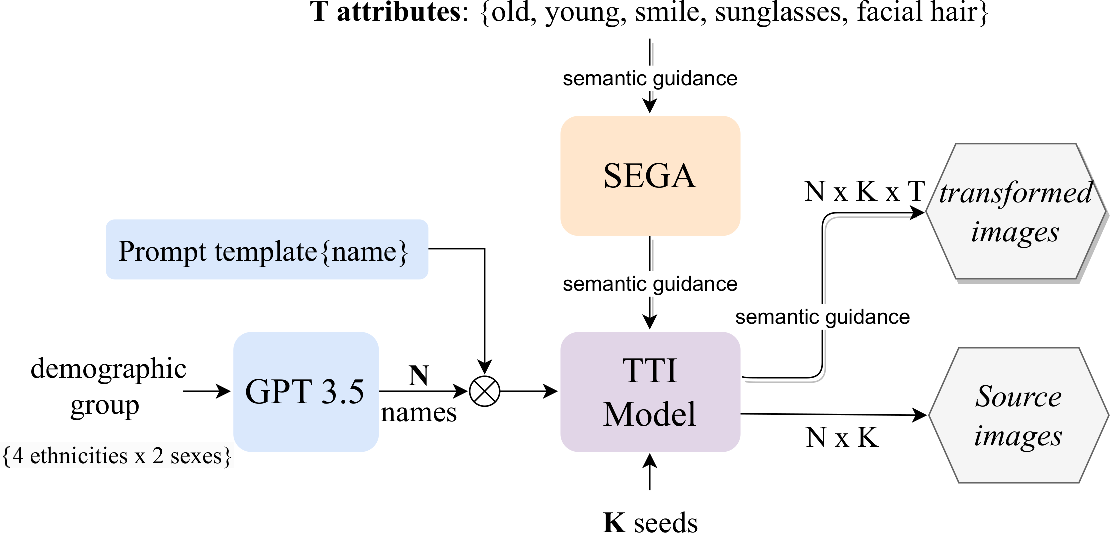}
    \caption{Our data generation pipeline.}
    \label{fig:pipeline}
\end{figure}

SDv2.1 is finetuned from the Stable Diffusion v2~(SDv2) checkpoint, which was trained from scratch on a subset of the LAION-5B dataset. SDv2.1's training dataset contains more faces than that of SDv2\footnote{\url{https://stability.ai/blog/stablediffusion2-1-release7-dec-2022}}. Hence, SDv2.1 performs better in generating faces than SDv2. Realism is among the many openly available fine-tuned models from the checkpoints of Stable Diffusion. However, its exact implementation details are not known. We treat both SDv2.1 and Realism as grey-box models. Both models are capable image generators with differing performance characteristics, and our framework is agnostic to their implementation details. The design of the system shown in~\cref{fig:pipeline} can be used with any relevant text-to-image generative model to synthesize scalable batches of facial data useful for training data augmentation or as tailored test sets for face recognition applications.

To diversify generated faces, we employ the semantic-guidance image generation technique SEGA~\cite{brack2023sega}. SEGA steers the TTI model towards generating images that incorporate semantic concepts based on user-provided textual edits while keeping the rest of the image semantics intact, all without the need for fine-tuning the TTI models. This technique proves valuable in creating faces with diverse attributes, such as incorporating sunglasses. 
Moreover, recent works~\cite{friedrich2023fair, smith2023balancing} leverage SEGA and similar methods for fair face image generation by introducing demographics as semantic concepts during image generation. Thus, we study the efficacy of incorporating SEGA in the image generation pipeline.

\subsection{Data Generation Pipeline}
\label{sec:gen_pipeline}
To generate our facial datasets, we design a prompt that specifies a demographic group and an identity associated with that group. The prompt guides the model to generate a set of diverse face images for each of these identities.

\paragraph{Identity.} We found that when we explicitly mention the demographic group in the prompt, like \emph{an Indian man}, the generated images exhibit limited diversity; i.e. identities look quite similar. To encourage the generation of more varied identities, we employed \emph{names} as indicators of different identities within demographic groups. We found that TTI models interpret names as proxies for ethnicity and sex, and each name carries a unique identity despite the randomness of the generation process. 

For each of the eight demographic groups we study in this paper, we instructed GPT-3.5 to create two separate lists of names—one comprising `celebrity' names and the other `non-celebrity' names. For the non-celebrity (celebrity) collection, we generated 20 (30) names per demographic group. The two lists reflect different levels of knowledge within the TTI model: celebrity images are more likely to exist in the training data of TTI, while non-celebrities are more likely to be indirectly learned by the model.

\paragraph{Prompt.} We desire prompts that guide the model to generate multiple and diverse face images with user-specified semantics. Including a name within a prompt encodes both identity and demographic information. Trial and error, combined with our user study, led us to the below approach. 

For Realism, we experimented with a set of prompts, and we found this template to generate face images of high quality: ``A photo of the face of $\{identity\}$.'' We vary the TTI generator seed to generate multiple images per identity and prompt. We also add a set of \textit{negative} prompts that steer the model away from unrealistic, cartoon, or low-quality image generation. These negative prompts are frequently used in face image generation. For a fair comparison, we use this prompt template along with a set of pre-selected five seeds to generate images of all identities and demographic groups.

For SDv2.1, we observed that the prior template generates images of poor quality on both celebrity and non-celebrity identities. Thus, we expanded the prompt template as follows: ``A photo of the face of ($\{identity\}$:2.0). (realistic:2.0). (Face shot only:2.0).'' This revised prompt improved the generated image quality of celebrity identities. However, it did not have the same effect on non-celebrity images. Thus, we decided to evaluate only the celebrity identities for the SDv2.1 model. 

We manually validated that the generated images from both models contain a face image, different seeds generate diverse images of the same identity, and that identities are distinct and belong to the intended demographic group. We provide further details on the challenges of high-quality face image generation in the long-form report.



\paragraph{Attributes.} 
Using SEGA with Realism and SDv2.1, we induce five attributes to the generated data: `young’, `old’, `facial hair’, `sunglasses’, and `smile.’ The details of SEGA's hyperparameters are in the long-form report. We refer to the images obtained without SEGA as \emph{source images} and with SEGA as \emph{transformed images}. All the synthesized images are of $512\times 512$ resolution. For SDv2.1, to ensure better quality, we generate the images at $768\times 768$ and then downsample them to $512\times 512$. Figure~\ref{fig:non-celeb-data} shows a sample of the non-celebrity images synthesized using Realism and SEGA.

In total, we generate 800 source and 4000 transformed non-celebrity images, and we generate 1200 source and 6000 transformed celebrity images per model.

\subsection{Evaluation Methods}
We use three independent evaluation methods to assess the quality of the generated datasets: quantitative metrics, face verification, and user study. 

\subsubsection{Quantitative Metrics} 

The metrics below are used to evaluate overall quality of the source and the transformed images.
\begin{itemize}
    \item \textbf{Image-Image Metrics:} These are mainly used to verify identity retention under SEGA transformation. CLIP-I and DINO-I measure the cosine similarity between the source and transformed images’ CLIP~\cite{radford2021learning} and  DINO-v2~\cite{oquab2023dinov2} embeddings, respectively. Higher similarity implies that the identity is preserved.
    
    \item \textbf{CLIP-directional}: CLIP-directional~\cite{gal2022stylegan} intends to identify the correctness of the semantic change in the transformed image. It measures the similarity of the change between the embeddings of the source and transformed images and the change between their captions.

\end{itemize}

\subsubsection{Face Verification}
Face verification accuracy utilizes pairwise face comparisons to measure embedding space quality. The embeddings of two faces depicting the same identity are expected to be close to each other. We analyze face verification performance on Facenet~\cite{schroff2015facenet}, a well-studied face recognition network.

Our analysis of face recognition models focuses on verification accuracy. Given two face images $\sample,\sample'$, verification accuracy $\ver$ is computed as:
\begin{align}
\begin{split}
\ver(\sample,\labelElement,\sample',\labelElement') 
&\definedAs \indicatorFunction[\labelElement = \labelElement']\cdot\indicatorFunction\big[\metric(\embeddingFunc_\embeddingDimension(\sample),\embeddingFunc_\embeddingDimension(\sample')) < \threshold \big] \\
& \;\; + \indicatorFunction[\labelElement \neq \labelElement']\cdot\indicatorFunction\big[\metric(\embeddingFunc_\embeddingDimension(\sample),\embeddingFunc_\embeddingDimension(\sample')) \geq \threshold \big]
\end{split}
\end{align}
where $\indicatorFunction$ denotes the indicator function and threshold $\threshold$ is chosen heuristically to minimize false verification rate. Further, $\labelElement$ and $\labelElement'$ are identities associated with $\sample$ and $\sample'$, respectively. We report the average verification accuracy as computed across sets of pairs. Within our evalutation, sets of pairs are constructed so that half of the pairs correspond to the same identity. When analyzing verification accuracy for the user study, we implicitly assume that humans can perfectly distinguish identities of generated faces.

To study the effect of demographics on verification, we report two notions of verification accuracy: same group and any group. For a specified group $\group$, same group verification accuracy refers to the evaluation of $\ver$ on lists of pairs in which both images $\sample,\sample'$ belong to group $\group$. Any group verification accuracy refers to the evaluation of $\ver$ where only each pair's first image $\sample$ must be in group $\group$. 

We utilize the Labeled Faces in the Wild (LFW) dataset as a baseline for natural faces verification. LFW is a canonical dataset for face recognition tasks. The LFW dataset contains 13233 images and a total of 5749 unique identities. Demographic annotations for images in LFW were obtained from the system introduced by Kumar et al.~\cite{kumar2009lfwattributes}.

\subsubsection{User Study} 
We conducted a human evaluation of the generated images from both models combined with SEGA. Toward that end, we designed an online Qualtrics survey for each model-identity collection pair, resulting in three surveys: (SDv2.1 Celebrities, Realism Celebrities, and Realism Non-Celebrities). The surveys are approved by our IRB and are conducted on the Prolific platform.

For each survey, we randomly sampled 15 identities per demographic group, one image per identity; 120 images in total. We paired each source image with its 5 transformed images corresponding to the 5 semantic attributes. This results in 600 source-transformed image pairs per survey. We presented each participant with a set of 21 blocks. Each block shows two images: one source image (without SEGA), and one transformed image (using SEGA), along with the transform instruction used by SEGA.  For each block, the participant answers three questions: (1) whether the two images depict the same person, (2) the consistency of the transformed image with the edit instruction on a 5-point scale, and (3) how they rate the quality of the two images on a 5-point Likert scale.

For each study, we recruited 85 participants, and each image pair received three ratings on average. Each participant was compensated \$3.5 for their effort, with an average completion time of 15 minutes. The study was distributed evenly to male and female participants. The participants' demographic distribution is discussed in our long-form report.

\section{Evaluation}\label{sec:Evaluation}

After generating the datasets, we apply the evaluation methods to analyze the associated demographic discrepancies.  Three questions guide this evaluation: 

    \begin{enumerate}
        \item \textit{How does face verification on synthetic data compare to natural data and does it exhibit demographic disparities?}
        \item \textit{Does the quality of synthetic face images depend on the demographic group? }
        \item \textit{Can quantitative metrics replace expensive user studies to assess the quality of synthetic face images? }
    \end{enumerate}


\begin{figure*}[t]
    \centering
    \begin{subfigure}[b]{\textwidth}
        \centering
        \includegraphics[width=0.75\textwidth]{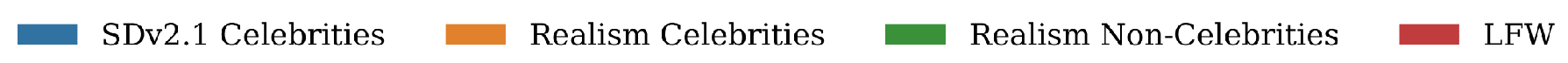} 
    \end{subfigure}
    \scalebox{0.95}{
    \begin{subfigure}[b]{0.32\textwidth}
        \includegraphics[width=\textwidth]{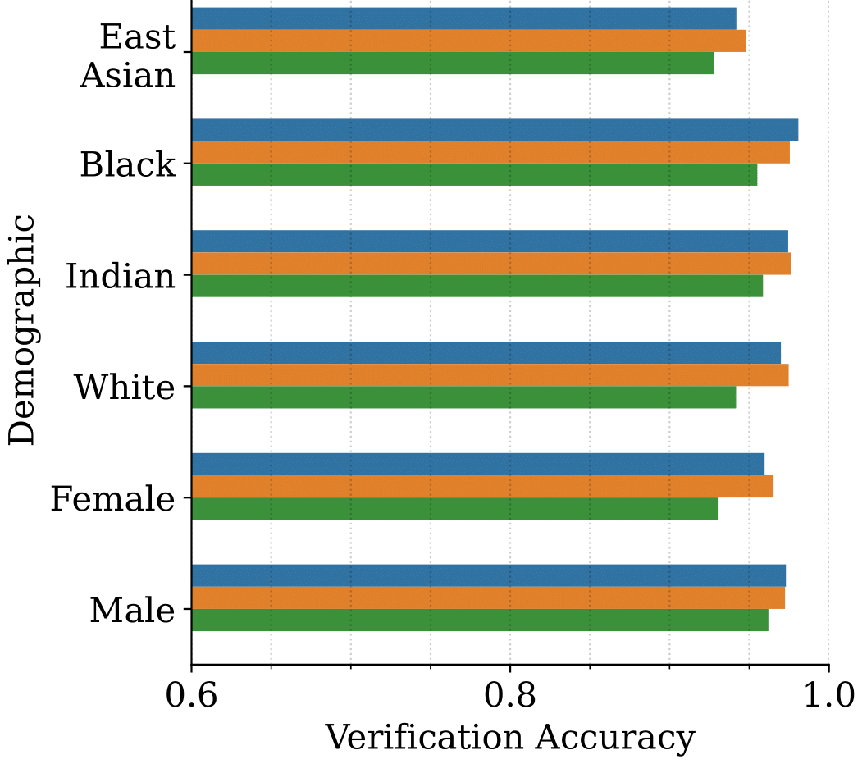}
        \caption{Survey Responses}
        \label{fig:image1}
    \end{subfigure}
    \hfill
    \begin{subfigure}[b]{0.32\textwidth}
        \includegraphics[width=\textwidth]{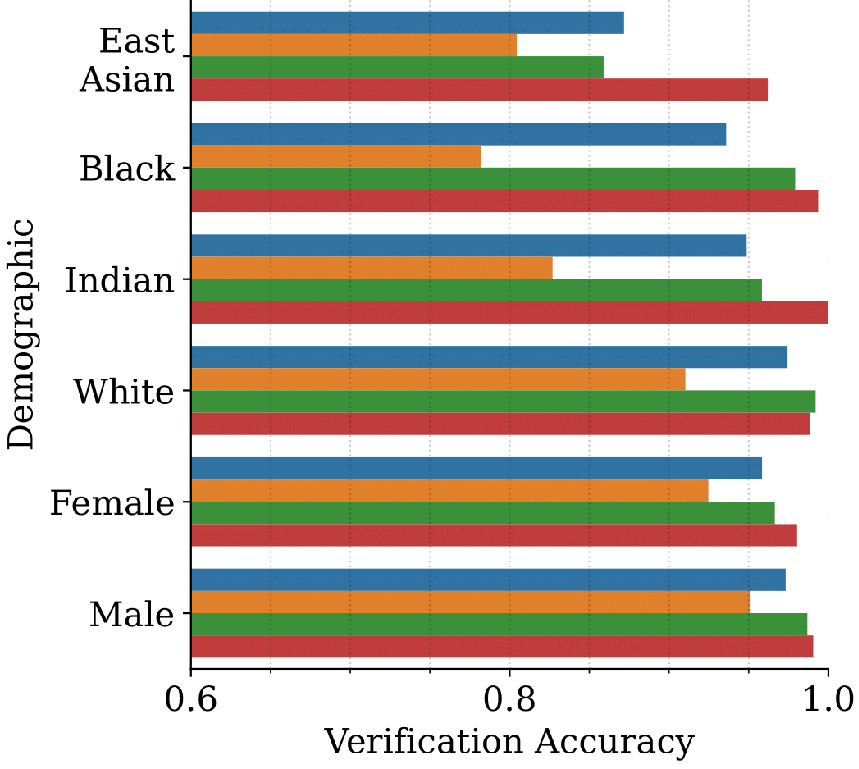}
        \caption{Same Demographic Pairs}
        \label{fig:image2}
    \end{subfigure}
    \hfill
    \begin{subfigure}[b]{0.32\textwidth}
        \includegraphics[width=\textwidth]{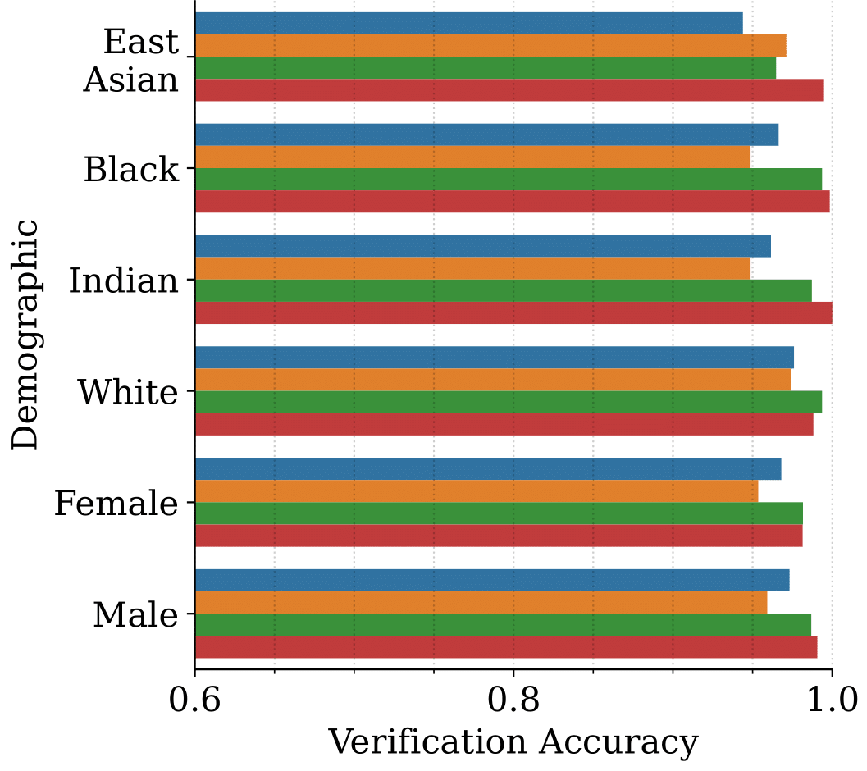}
        \caption{Any Demographic Pairs}
        \label{fig:image3}
    \end{subfigure}}
    \caption{Verification Accuracy is plotted across four datasets. Each row is a demographic, and each dataset is depicted with a different hue. Note that each plot is x-axis limited between 0.6 and 1.}
    \label{fig:verification}
\end{figure*}

\subsection{Face Verification}

Face verification performance is depicted in \cref{fig:verification}. The figure shows the verification accuracy measured on LFW and synthetic datasets. Across all demographics and datasets, with one exception, we observe that generated faces perform worse than natural faces (LFW). Only in the White demographic does a synthetic dataset, Realism Celebrities, have better face verification performance than natural data. We also observe that for each demographic and dataset pair, same-demographic verification accuracy is often notably less than its any-demographic counterpart. Hence, we conclude that face recognition systems are demographically aware on generated faces in a manner similar to natural faces.





\subsection{Synthetic Face Image Quality}

\begin{table}[t]
\small
    \centering
    \resizebox{\columnwidth}{!}{
    \begin{tabular}{l c cccccc }
    \toprule
     & \multirow{2}{*}{\textbf{Dataset}} & \multicolumn{6}{c}{\textbf{Demographic group}} \\
    & & E Asian & Black & Indian & White & Female & Male \\
    \midrule
    \multirow{3}{*}{\textbf{$\mathsf{M1}$}} & D1 & 4.463 & 4.401 & \textbf{4.394} & \textbf{4.515} & 4.428 & 4.459 \\
    & D2 &  \textbf{4.253} & 4.240 & \textbf{4.045} & 4.097 & 4.166 & 4.140 \\
    & D3 & 4.121 & \textbf{4.149 }& 4.112 & \textbf{4.039} & 4.112 & 4.099 \\
    \midrule
    \multirow{3}{*}{\textbf{$\mathsf{M2}$}} & D1 & 0.195 & 0.122 & 0.187 & 0.184 & \textbf{0.244} & \textbf{0.098} \\
    & D2 & 0.190 & \textbf{0.085} & 0.114 & \textbf{0.592} & 0.364 & 0.140 \\
    & D3 & 0.100 & \textbf{0.156} & 0.123 & 0.123 & 0.154 & \textbf{0.098} \\
    \midrule
    \multirow{3}{*}{\textbf{$\mathsf{M3}$}} & D1 & 4.020 & 3.972 & \textbf{4.144} & \textbf{3.945} & 3.954 & 4.092 \\
    & D2 & 3.624 & 3.367 & \textbf{3.717} & \textbf{2.636} & 3.203 & 3.455 \\
    & D3 & \textbf{3.747} & \textbf{3.197} & 3.516 & 3.325 & 3.492 & 3.412 \\
    \midrule
    \multirow{3}{*}{\textbf{$\mathsf{M4}$}} & D1 &     87.5 & 84.6 & \textbf{90.3} & \textbf{83.8} & 85.5 & 87.7 \\
    & D2 & 78.4 & 69.5 & \textbf{80.7} & \textbf{51.2} & 66.0 & 73.6 \\
    & D3 & \textbf{79.9} & \textbf{65.2} & 72.3 & 69.9 & 72.2 & 71.8 \\
    \bottomrule
    \end{tabular}}
    \caption{User survey average answers to the following measures: $\mathsf{M1}$: source image quality on a 5-point scale, $\mathsf{M2}$: drop in image quality after SEGA transformation, $\mathsf{M3}$: SEGA transformation correctness on a 5-point scale, and $\mathsf{M4}$: percentage (\%) of correct transformation (transformation correctness score $\geq$ 3 out of 5). {D1}: Realism Non-Celebrities, {D2}: Realism Celebrities, {D3}: SDv2.1 Celebrities. Highest and lowest scores are highlighted in \textbf{bold}. E Asian denotes the East Asian demographic group.}
    \label{tab:study_answers}
\end{table}

Table~\ref{tab:study_answers} presents the average survey scores in terms of image quality and transformation correctness across all demographics and datasets. The scores suggest that image quality depends on the identity's demographic group. Moreover, SEGA transformations drop the quality of all images, and the drop is also demographic-dependent. We use one-way ANOVA in an attempt to reject null hypotheses of forms: 
\begin{nullhypothesis}\label{nullhyp:sourcequality}
    The per-demographic distributions of source image quality in \textlangle Dataset\textrangle~are identical.
\end{nullhypothesis}

\begin{nullhypothesis}\label{nullhyp:targetquality}
    The per-demographic distributions of transformed image quality in \textlangle Dataset\textrangle~are identical.
\end{nullhypothesis}

\begin{nullhypothesis}\label{nullhyp:sourcetargetdrop}
    The per-demographic distributions of quality difference between source and transformed images in \textlangle Dataset\textrangle~are identical.
\end{nullhypothesis}



 On the Realism Non-Celebrities and Realism Celebrities datasets, one-way ANOVA rejects  \cref{nullhyp:sourcequality,nullhyp:targetquality,nullhyp:sourcetargetdrop} with $p$-values less than 0.05; corresponding $p$-values appear in the long-form report. This test tells us that for these two datasets, source image quality, transformed image quality, and the difference between source and transformed image quality have a dependence on demographics.
 The only dataset for which image quality does not conclusively depend on demographics is SDv2.1 Celebrities.

The same observation of demographic dependence applies to the transformation correctness measures ($\mathsf{M3}$, $\mathsf{M4}$). It is interesting to note that demographic groups that have higher source image quality are not consistent with groups of higher transformation correctness. This suggests that SEGA introduces its own biases in the generative pipeline. 

\subsection{Quantitative Metrics vs. User Study}

\begin{table}[t]
\centering

\label{tab:pvalues}
\resizebox{\columnwidth}{!}{
\begin{tabular}{lccc}
\toprule
\multirow{2}{*}{\textbf{Dataset}} & \multicolumn{2}{c}{\textbf{\Cref{nullhyp:imagesimilariymetricCorrelation}}} & \textbf{\Cref{nullhyp:directionalsimilaritymetricCorrelation}}\\
\cmidrule{2-4}
& \textbf{CLIP-I} & \textbf{DINO-I} & \textbf{CLIP Directional}\\
\midrule

SDv2.1 Celebrities 
 & $0.147$ & $0.107$ & $0.128$ \\
\midrule
Realism Celebrities 
& $0.197$ & $0.122$ & $0.348$\\
\midrule
Realism Non-Celebrities 
& $0.245$ & $0.142$ & $0.0908$ \\
\bottomrule
\end{tabular}}
\caption{Spearman correlation coefficients for \cref{nullhyp:imagesimilariymetricCorrelation,nullhyp:directionalsimilaritymetricCorrelation}. Each correlation coefficient is statistically significant. Corresponding $p$-values appear in the long-form report.}\label{tab:spearmanmetrics}
\end{table}

User studies are the most direct way to measure human perception of generated faces. Unfortunately, they are prohibitively expensive when implemented at scale. If we have a metric serving as a proxy for human sentiment toward generated face quality, costs associated with generating realistic face data could be drastically reduced. We analyze the correlation between the different metrics and the questions posed in the user study regarding the quality of the source and transformed images, the presence of semantic change, and identity retention after applying the semantic change. We calculate the Spearman correlation coefficients between the metrics and the scores to the user-study questions and once again make use of one-way ANOVA tests to reject null hypotheses:

\begin{nullhypothesis}\label{nullhyp:imagesimilariymetricCorrelation}
    On \textlangle Dataset\textrangle, there is no monotonic relationship between image-image \textlangle similarity metric\textrangle~and maintenance of identity post application of semantic change.
\end{nullhypothesis}

\begin{nullhypothesis}\label{nullhyp:directionalsimilaritymetricCorrelation}
    On \textlangle Dataset\textrangle, there is no monotonic relationship between CLIP-directional and appearance of the semantic change.
\end{nullhypothesis}

On all three datasets, one-way ANOVA tests enable us to reject \cref{nullhyp:imagesimilariymetricCorrelation} on image similarity metrics CLIP-I and DINO-I. We also similarly reject \cref{nullhyp:directionalsimilaritymetricCorrelation} on the CLIP Directional metric. Despite rejecting null hypotheses, each Spearman coefficient is low, as evident from~\cref{tab:spearmanmetrics}. Hence, in the context of face recognition, image quality metrics are not a suitable proxy for humans in performing both identity verification and transformation verification tasks. This result is partially surprising: the DINO-I metric is designed to recognize differences between images of similar description. However, CLIP-I metric exhibits difficulty in distinguishing images with similar text descriptions~\cite{ruiz2023dreambooth}. Our findings also indicate a low correlation between this metric and human assessment.

\section{Analytical Model}
We observed that verification accuracy degrades on synthetic images. We attribute this to machine learning models being trained on finite data. Typically these datasets are drawn from the internet. Generative models, such as diffusion models, thusly learn to generate images patterned on their internet-based dataset. Because the internet is well-understood to be a biased sample of the universe,
a diffusion network trained on an internet-sourced dataset is itself a biased sample generator. To understand how a biased, finite training set can yield biased sample generation, we utilize a Gaussian Mixture Model (GMM). A GMM is theoretically tractable proxy through which we gain intuition about generative models. 

In that model, an image in demographic group $\group$ is drawn from $\normalDist(\vec{\mu}_\group,\vec{\Sigma}_\group)$. For brevity, we denote the distribution of examples in group $\group$ by $\sampleDist_\group$. Without loss of generality, our analysis considers two groups: $a$ and $b$. Group $a$ occurs with probability $\alpha$ where $\alpha \in (0,1)$. Thus, group $b$ occurs with probability $1-\alpha$. The universe's distribution can be written as $\sampleDist = \alpha \sampleDist_a + (1-\alpha)\sampleDist_b$. As previously identified, generative models are typically trained on a biased dataset. To model this bias, we assume training dataset $\trainingData$ is drawn i.i.d. from biased data distribution $\sampleDist_\trainingData$. Samples in $\trainingData$ are assumed to be $\dimension$-dimensional.  The biased data distribution $\sampleDist_\trainingData$ is a possibly re-weighted mixture of Gaussians $\sampleDist_a$ and $\sampleDist_b$. That is to say, $\sampleDist_\trainingData = \beta\sampleDist_a + (1-\beta)\sampleDist_b$ where $\beta \in (0,1)$. Distributions $\sampleDist_\trainingData$ and $\sampleDist$ are only equivalent if $\alpha = \beta$. For notational brevity, $\trainingData_\group$ denotes examples in $\trainingData$ drawn from $\sampleDist_\group$. 

The estimator of $\sampleDist$ learned from $\trainingData$ is denoted $\hat{\sampleDist}_\trainingData$. The quality of estimator $\hat{\sampleDist}_\trainingData$ is measured with total variation distance, a notion of distributional discrepancy. Our use of total variation distance as a notion of distribution estimator quality is motivated by its use in generative model literature~\cite{lin2018pacgan,sajjadi2018assessing}.

\begin{definition}[Discrepancy]\label{def:discrepancy}
Consider a measure space $(\Omega,\mathcal{F})$. If $\placeholderDist_1$ and $\placeholderDist_2$ are continuous probability  distributions, then the discrepancy is computed as 
\begin{equation}
    \metric_{\totalVariation}(\placeholderDist_1,\placeholderDist_2) = \sup_{\placeholder \in \mathcal{F}}\abs{\placeholderDist_1(\placeholder)-\placeholderDist_2(\placeholder)}
\end{equation}
\end{definition}

Utilizing \cref{def:discrepancy}, we can show that two distinct factors contribute to discrepancy. The first factor is the finite size of $\trainingData$.  The second factor is $\trainingData$ being a non-representative sample of $\sampleDist$. Both factors are formalized in \cref{prop:wassersteinDiscrepancy}, its proof is in our long-form report. We assume the process yielding $\hat{\sampleDist}_\trainingData$ is an empirical Bayes estimator, such as the Expectation-Maximization algorithm. This process learns five parameters about the distribution: the group proportion $\beta$, the means and covariances of groups $a$ and $b$: $\vec{\mu}_{\groupA}$, $\vec{\Sigma}_\groupA$, $\vec{\mu}_{\groupB}$, and $\vec{\Sigma}_\groupB$.

\begin{proposition}[Proposition]\label{prop:wassersteinDiscrepancy}
Let $\delta \in (0,1)$. If 

\begin{equation}
\small
\abs{\trainingData} = \bigOh\bigg(\dimension^2\log(2) \log (2/\delta)\bigg[H^2(\sampleDist_a,\sampleDist_b)\bigg]^{-4}\bigg) 
\end{equation}
then we have 
\begin{equation}
\small
\Pr\bigg[\metric_{\totalVariation}(\hat{\sampleDist}_\trainingData,\sampleDist) > \frac{\abs{\alpha-\beta}}{2}H^2(\sampleDist_a,\sampleDist_b)\bigg] \geq 1  - \delta
\end{equation}
where $H^2$ is the squared Hellinger distance, and 

\begin{equation}
\small
\begin{split}
\small
 & H^2(\sampleDist_a,\sampleDist_b) = 
 \bigg(1-\frac{\abs{\vec{\Sigma}_a}^{\sfrac{1}{4}}\abs{\vec{\Sigma}_b}^{\sfrac{1}{4}} }{\abs{\frac{\vec{\Sigma}_a+\vec{\Sigma}_b}{2}}^{\sfrac{1}{2}}}\bigg)  \\
& \times \exp\bigg\{-\frac{1}{8}(\vec{\mu}_a-\vec{\mu}_b)^\transpose\bigg(\frac{\vec{\Sigma}_a+\vec{\Sigma}_b}{2}\bigg)^{-1}(\vec{\mu}_a-\vec{\mu}_b)\bigg\}
\end{split}
\end{equation} 

if $\sampleDist_a,\sampleDist_b$ are each multivariate Gaussians.
\end{proposition}

This proposition suggests that even with a large number of samples in $\trainingData$, it can be impossible to learn $\sampleDist$ exactly. This is due to non-representative proportions being drawn from each demographic group, i.e. when $\abs{\alpha-\beta}>0$. On the other hand, when $\alpha=\beta$, and the number of samples in $\trainingData$ is infinite, $\hat{\sampleDist}_\trainingData$ and $\sampleDist$ are equal, so $\metric_{\totalVariation}(\hat{\sampleDist}_\trainingData,\sampleDist)$ tends to $0$. The bound also has a dependence on dimension squared: large dimension inputs require many more samples to train high-fidelity generative models. Thus, we conjecture faces generated by diffusion models trained upon larger datasets should close the observed gap in verification accuracy between LFW and datasets synthesized in this paper.

\subsection{Empirical Validation}
\begin{figure}
    \centering
    \begin{subfigure}{0.45\columnwidth}
        \centering
        \includegraphics[width=\columnwidth]{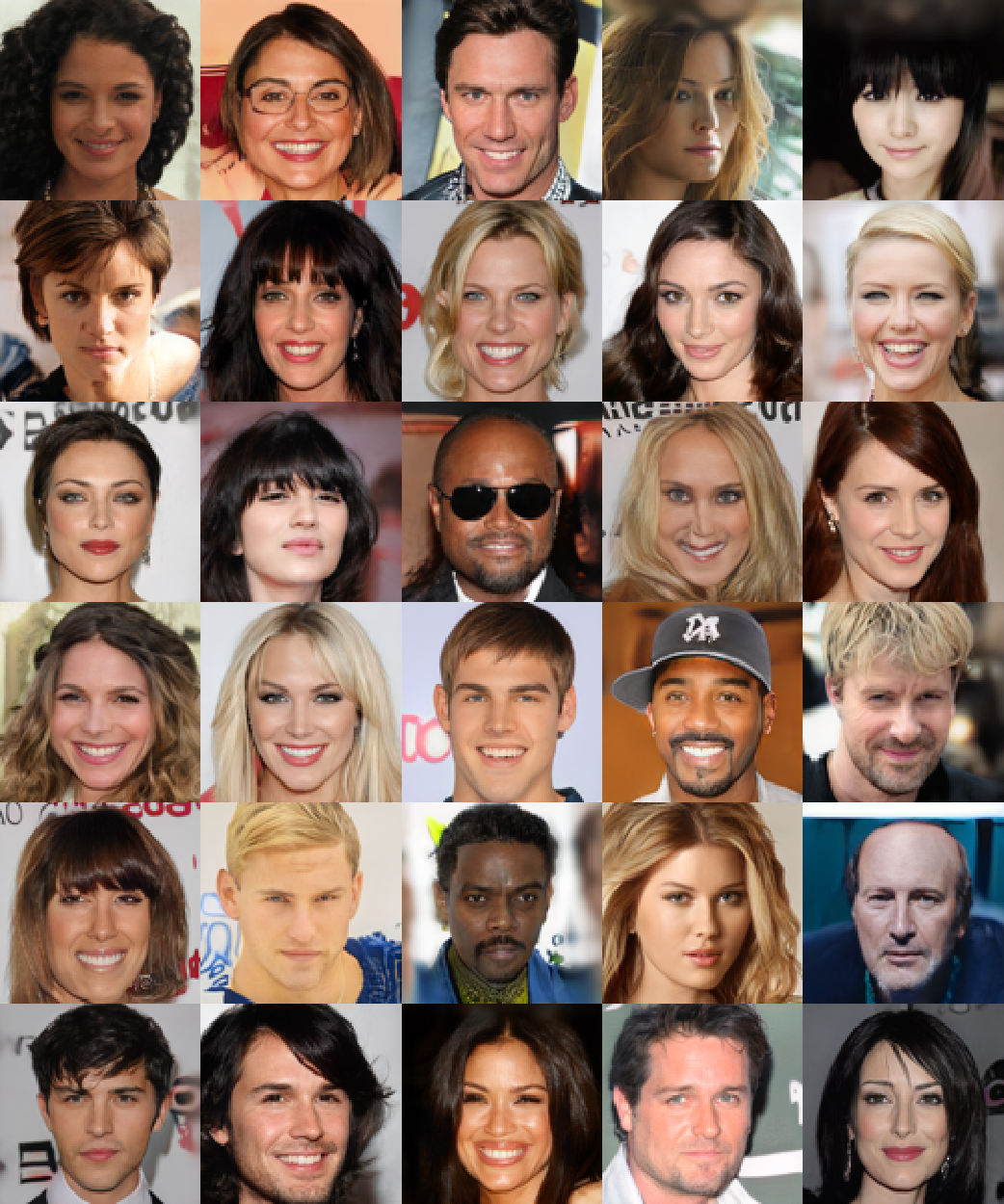}
        \caption{Inliers: mostly female and higher perceptual quality}
    \end{subfigure}%
    \hfill
    \begin{subfigure}{0.45\columnwidth}
        \centering
        \includegraphics[width=\columnwidth]{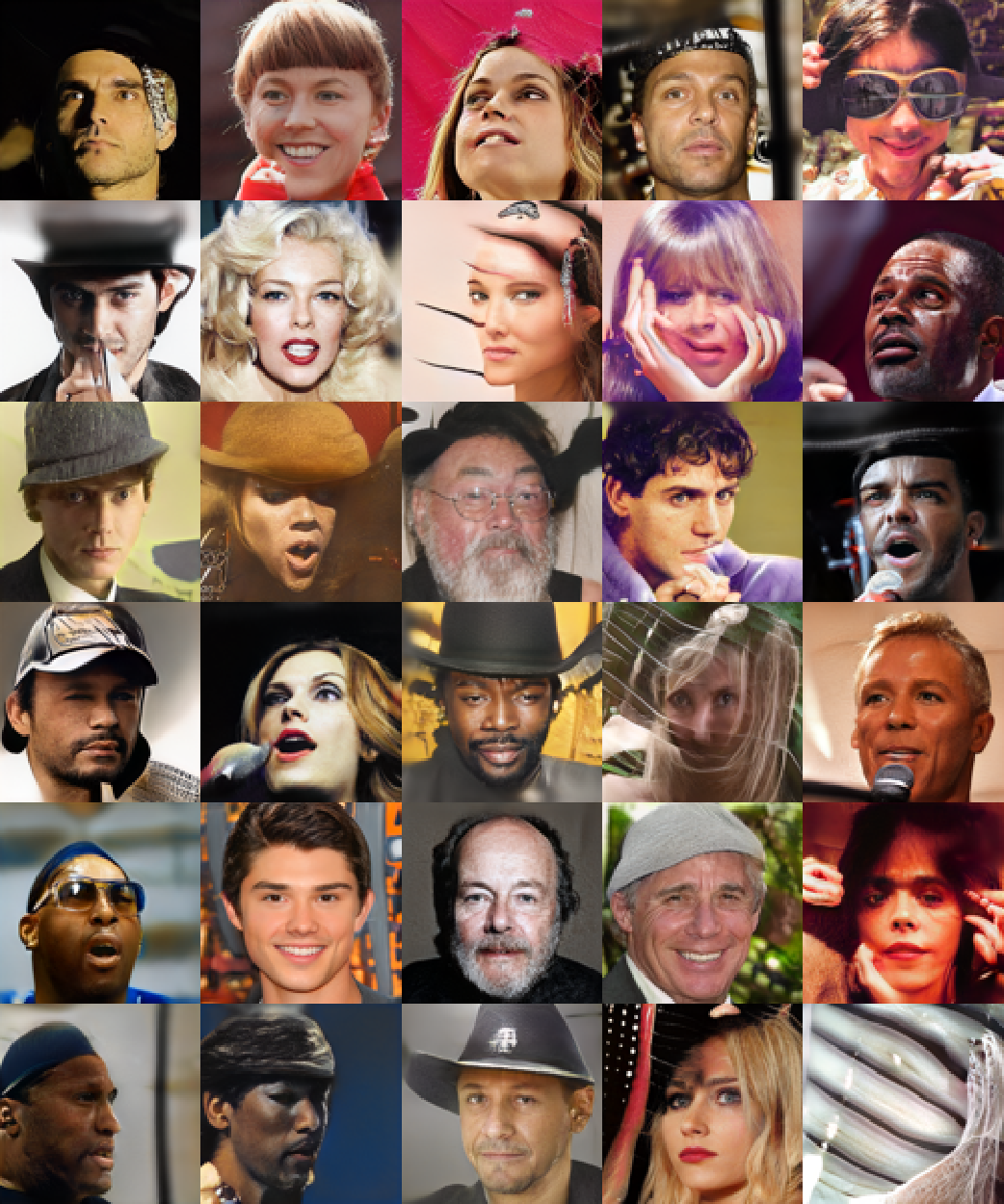}
        \caption{Outliers: mostly male and lower perceptual quality}
    \end{subfigure}
    \caption{Images generated from StyleGAN trained with CelebA~(female-biased). Outliers and Inliers obtained using 5-nearest neighbor approach.}\label{fig:example_fidelity}
\end{figure}

We further to validate our theoretical model using a strategy similar to~\citeauthor{naeem2020reliable}. In particular, we generate and assess the quality of faces from a StyleGAN trained on CelebA~\cite{liu2015faceattributes}, an open-source face dataset. The CelebA training set is majority (over 58\%) female, hence our theory suggests female examples will have better fidelity to ground truth examples. We follow \citeauthor{naeem2020reliable}'s method to evaluate the quality of faces generated by StyleGAN: we observe that the faces with higher quality are those of females. \Cref{fig:example_fidelity} shows examples of the generated faces broken into two categories: inliers (those with high fidelity) that are mostly female and outliers (those with low fidelity) that are mostly male.

\begin{figure}[t]
    \centering
    \begin{subfigure}[b]{\columnwidth}
        \centering
        \includegraphics[width=0.9\columnwidth]{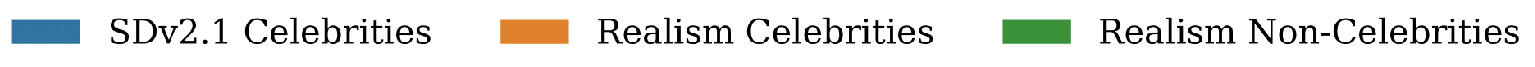} 
    \end{subfigure}
    \resizebox{\columnwidth}{!}{
    
    \begin{subfigure}[b]{0.25\textwidth}
        \includegraphics[width=\textwidth]{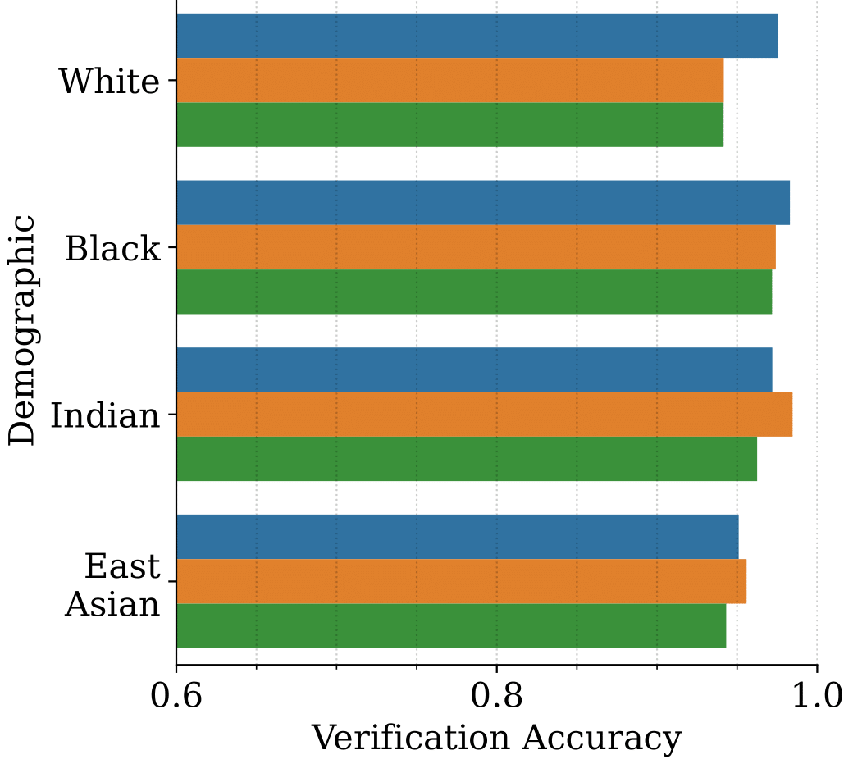}
        \caption{Black}
        \label{fig:image11}
    \end{subfigure}
    \hfill
    \begin{subfigure}[b]{0.25\textwidth}
        \includegraphics[width=\textwidth]{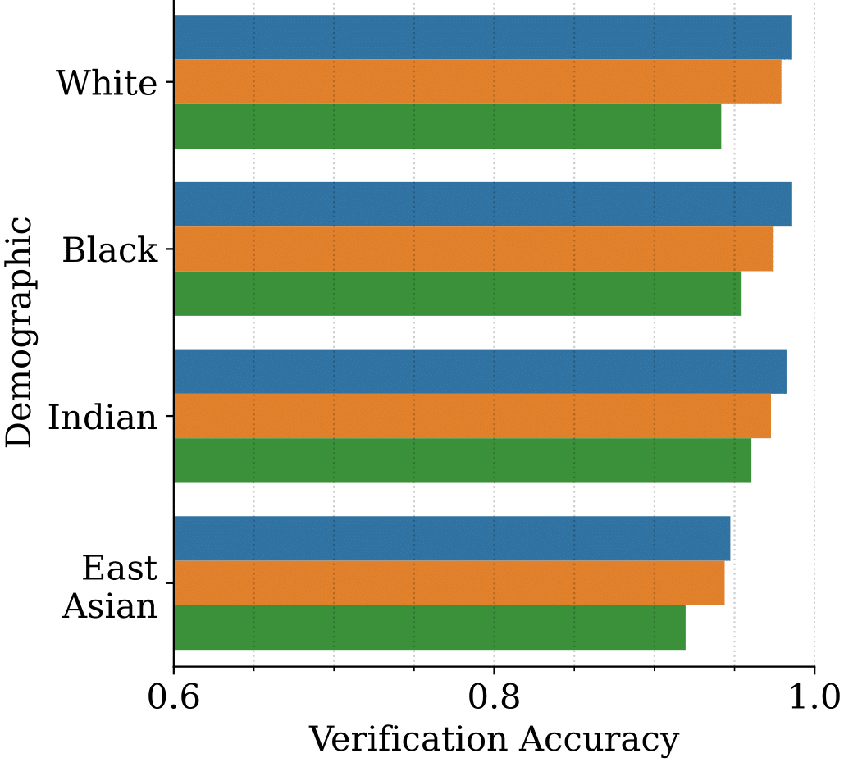}
        \caption{White}
        \label{fig:image22}
    \end{subfigure}
    \hfill
    \begin{subfigure}[b]{0.25\textwidth}
        \includegraphics[width=\textwidth]{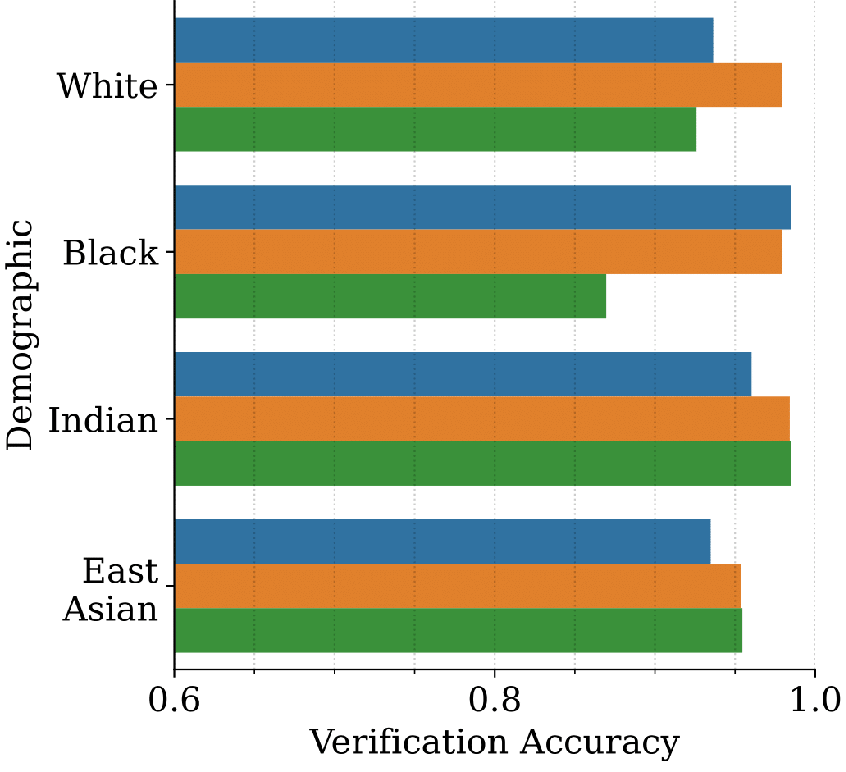}
        \caption{Hispanic}
        \label{fig:image33}
    \end{subfigure}
    }
    \caption{User Verification Accuracy. The $y$-axis captures queried image demographics. Each subfigure depicts a respondent demographic. Note that each plot is x-axis limited between 0.6 and 1.}
    \label{fig:respondent_demographic_verification}
\end{figure}

\section{Discussion}
Our user study provides direction for follow-on research relating to the Own Race Effect (ORE). ORE refers to the documented tendency of individuals to better recognize faces from within their racial group~\cite{TANAKA2004B1,meissner2001thirty}. We observed that our user survey seems to disagree with ORE as shown in \cref{fig:respondent_demographic_verification}. Hence, a rigorous study of perceived identity of images under semantic transformations would be of research value.

As evidenced by the user study, mechanisms of human face perception present unique challenges to the application of generative models in face recognition. Moreover, automated prompt design strategies require access to a metric quantifying the quality of generated images. This does not detract from techniques evaluating generative model performance, rather, it opens a new research avenue: tuning face quality metrics to better align with human preferences.

Though our analysis techniques generalize to other domains, they assume CLIP functions as intended. Unfortunately, CLIP and similar semantic-visual embedding models are trained on internet data. Hence, their embedding space contains biases. Further, CLIP is known to have trouble constructing embeddings for uncommon or otherwise niche words and phrases. Niche words and phrases, such as ``inter-eye distance'' and ``eyebrow slant'', which can affect human perceived face identity~\cite{tsao2008mechanisms}, are problematic for CLIP. Further analysis of semantic-visual embeddings is necessary to gain a full picture of text-to-image generative models. Additionally, CLIP's understanding of cultural constructs is not entirely understood. For example, it is unclear what an ``intelligent face'' or ``beautiful face'' means to CLIP. Thus, the semantic transformations we study are explicit face attributes.

Finally, our study does not negate the value of generative approaches in model analysis. Generative examples can serve as a targeted curated dataset. Tailored generation has the potential to mitigate inherent biases found in existing datasets; however, the effectiveness of this approach is closely tied to the data used to train the generative model. It's important to remember that the generated examples are not i.i.d. samples from the natural distribution. Instead, they represent i.i.d. samples from a possibly skewed estimate derived from a finite pool of realized examples within the training set.

\section{Conclusion}

Generative models have been the subject of much recent societal interest. Synthesized examples achieve near-realistic quality. Though recent advances have increased the expressive power of generative models, their performance characteristics remain opaque, especially for face image generation. We put forth a new framework to synthesize diverse face images and evaluate them from multiple perspectives. Our findings are boosted with intuition from an analytical model. Our work demonstrates the need for further research into properties of semantic-visual embeddings and human perception mechanisms upon generated faces. 
\section{Acknowledgments}

This work is partially supported by the DARPA GARD program under agreement number 885000, the NSF through award CNS-1942014, and the Wisconsin Alumni Research Foundation. The authors would also like to thank Yue Gao\footnote{gy@cs.wisc.edu} for his thoughtful insights that led to the research in this work.

\bibliographystyle{plainnat}
\bibliography{main}

\begin{thebibliography}{33}
\providecommand{\natexlab}[1]{#1}
\providecommand{\url}[1]{\texttt{#1}}
\expandafter\ifx\csname urlstyle\endcsname\relax
  \providecommand{\doi}[1]{doi: #1}\else
  \providecommand{\doi}{doi: \begingroup \urlstyle{rm}\Url}\fi

\bibitem[Ashtiani et~al.(2018)Ashtiani, Ben-David, and Mehrabian]{ashtiani2018sample}
Hassan Ashtiani, Shai Ben-David, and Abbas Mehrabian.
\newblock Sample-efficient learning of mixtures.
\newblock In \emph{Proceedings of the AAAI Conference on Artificial Intelligence}, volume~32, 2018.

\bibitem[Brack et~al.(2023)Brack, Friedrich, Hintersdorf, Struppek, Schramowski, and Kersting]{brack2023sega}
Manuel Brack, Felix Friedrich, Dominik Hintersdorf, Lukas Struppek, Patrick Schramowski, and Kristian Kersting.
\newblock Sega: Instructing diffusion using semantic dimensions.
\newblock \emph{arXiv preprint arXiv:2301.12247}, 2023.

\bibitem[Brooks et~al.(2023)Brooks, Holynski, and Efros]{brooks2022instructpix2pix}
Tim Brooks, Aleksander Holynski, and Alexei~A. Efros.
\newblock Instructpix2pix: Learning to follow image editing instructions.
\newblock In \emph{CVPR}, 2023.

\bibitem[Buolamwini and Gebru(2018)]{buolamwini2018gender}
Joy Buolamwini and Timnit Gebru.
\newblock Gender shades: Intersectional accuracy disparities in commercial gender classification.
\newblock In \emph{Conference on fairness, accountability and transparency}, pages 77--91. PMLR, 2018.

\bibitem[Dixit et~al.(2017)Dixit, Kwitt, Niethammer, and Vasconcelos]{dixit2017aga}
Mandar Dixit, Roland Kwitt, Marc Niethammer, and Nuno Vasconcelos.
\newblock Aga: Attribute-guided augmentation.
\newblock In \emph{Proceedings of the IEEE Conference on Computer Vision and Pattern Recognition}, pages 7455--7463, 2017.

\bibitem[Friedrich et~al.(2023)Friedrich, Schramowski, Brack, Struppek, Hintersdorf, Luccioni, and Kersting]{friedrich2023fair}
Felix Friedrich, Patrick Schramowski, Manuel Brack, Lukas Struppek, Dominik Hintersdorf, Sasha Luccioni, and Kristian Kersting.
\newblock Fair diffusion: Instructing text-to-image generation models on fairness.
\newblock \emph{arXiv preprint arXiv:2302.10893}, 2023.

\bibitem[Gal et~al.(2022{\natexlab{a}})Gal, Alaluf, Atzmon, Patashnik, Bermano, Chechik, and Cohen-Or]{gal2022image}
Rinon Gal, Yuval Alaluf, Yuval Atzmon, Or~Patashnik, Amit~H Bermano, Gal Chechik, and Daniel Cohen-Or.
\newblock An image is worth one word: Personalizing text-to-image generation using textual inversion.
\newblock \emph{arXiv preprint arXiv:2208.01618}, 2022{\natexlab{a}}.

\bibitem[Gal et~al.(2022{\natexlab{b}})Gal, Patashnik, Maron, Bermano, Chechik, and Cohen-Or]{gal2022stylegan}
Rinon Gal, Or~Patashnik, Haggai Maron, Amit~H Bermano, Gal Chechik, and Daniel Cohen-Or.
\newblock Stylegan-nada: Clip-guided domain adaptation of image generators.
\newblock \emph{ACM Transactions on Graphics (TOG)}, 41\penalty0 (4):\penalty0 1--13, 2022{\natexlab{b}}.

\bibitem[Kawar et~al.(2023)Kawar, Zada, Lang, Tov, Chang, Dekel, Mosseri, and Irani]{kawar2023imagic}
Bahjat Kawar, Shiran Zada, Oran Lang, Omer Tov, Huiwen Chang, Tali Dekel, Inbar Mosseri, and Michal Irani.
\newblock Imagic: Text-based real image editing with diffusion models.
\newblock In \emph{Proceedings of the IEEE/CVF Conference on Computer Vision and Pattern Recognition}, pages 6007--6017, 2023.

\bibitem[Kumar et~al.(2009)Kumar, Berg, Belhumeur, and Nayar]{kumar2009lfwattributes}
N.~Kumar, A.~C. Berg, P.~N. Belhumeur, and S.~K. Nayar.
\newblock {A}ttribute and {S}imile {C}lassifiers for {F}ace {V}erification.
\newblock In \emph{IEEE International Conference on Computer Vision (ICCV)}, Oct 2009.

\bibitem[Lin et~al.(2018)Lin, Khetan, Fanti, and Oh]{lin2018pacgan}
Zinan Lin, Ashish Khetan, Giulia Fanti, and Sewoong Oh.
\newblock Pacgan: The power of two samples in generative adversarial networks.
\newblock \emph{Advances in neural information processing systems}, 31, 2018.

\bibitem[Liu et~al.(2015)Liu, Luo, Wang, and Tang]{liu2015faceattributes}
Ziwei Liu, Ping Luo, Xiaogang Wang, and Xiaoou Tang.
\newblock Deep learning face attributes in the wild.
\newblock In \emph{Proceedings of International Conference on Computer Vision (ICCV)}, December 2015.

\bibitem[Luccioni et~al.(2023)Luccioni, Akiki, Mitchell, and Jernite]{luccioni2023stable}
Alexandra~Sasha Luccioni, Christopher Akiki, Margaret Mitchell, and Yacine Jernite.
\newblock Stable bias: Analyzing societal representations in diffusion models.
\newblock \emph{arXiv preprint arXiv:2303.11408}, 2023.

\bibitem[Maluleke et~al.(2022)Maluleke, Thakkar, Brooks, Weber, Darrell, Efros, Kanazawa, and Guillory]{maluleke2022studying}
Vongani~H Maluleke, Neerja Thakkar, Tim Brooks, Ethan Weber, Trevor Darrell, Alexei~A Efros, Angjoo Kanazawa, and Devin Guillory.
\newblock Studying bias in gans through the lens of race.
\newblock In \emph{European Conference on Computer Vision}, pages 344--360. Springer, 2022.

\bibitem[Meissner and Brigham(2001)]{meissner2001thirty}
Christian~A Meissner and John~C Brigham.
\newblock Thirty years of investigating the own-race bias in memory for faces: A meta-analytic review.
\newblock \emph{Psychology, Public Policy, and Law}, 7\penalty0 (1):\penalty0 3, 2001.

\bibitem[Mu{\~n}oz et~al.(2023)Mu{\~n}oz, Zannone, Mohammed, and Koshiyama]{munoz2023uncovering}
Cristian Mu{\~n}oz, Sara Zannone, Umar Mohammed, and Adriano Koshiyama.
\newblock Uncovering bias in face generation models.
\newblock \emph{arXiv preprint arXiv:2302.11562}, 2023.

\bibitem[Naeem et~al.(2020)Naeem, Oh, Uh, Choi, and Yoo]{naeem2020reliable}
Muhammad~Ferjad Naeem, Seong~Joon Oh, Youngjung Uh, Yunjey Choi, and Jaejun Yoo.
\newblock Reliable fidelity and diversity metrics for generative models.
\newblock In \emph{International Conference on Machine Learning}, pages 7176--7185. PMLR, 2020.

\bibitem[Oquab et~al.(2023)Oquab, Darcet, Moutakanni, Vo, Szafraniec, Khalidov, Fernandez, Haziza, Massa, El-Nouby, Assran, Ballas, Galuba, Howes, Huang, Li, Misra, Rabbat, Sharma, Synnaeve, Xu, Jegou, Mairal, Labatut, Joulin, and Bojanowski]{oquab2023dinov2}
Maxime Oquab, Timothée Darcet, Théo Moutakanni, Huy Vo, Marc Szafraniec, Vasil Khalidov, Pierre Fernandez, Daniel Haziza, Francisco Massa, Alaaeldin El-Nouby, Mahmoud Assran, Nicolas Ballas, Wojciech Galuba, Russell Howes, Po-Yao Huang, Shang-Wen Li, Ishan Misra, Michael Rabbat, Vasu Sharma, Gabriel Synnaeve, Hu~Xu, Hervé Jegou, Julien Mairal, Patrick Labatut, Armand Joulin, and Piotr Bojanowski.
\newblock Dinov2: Learning robust visual features without supervision, 2023.

\bibitem[Perera and Patel(2023)]{perera2023analyzing}
Malsha~V Perera and Vishal~M Patel.
\newblock Analyzing bias in diffusion-based face generation models.
\newblock \emph{arXiv preprint arXiv:2305.06402}, 2023.

\bibitem[Radford et~al.(2021)Radford, Kim, Hallacy, Ramesh, Goh, Agarwal, Sastry, Askell, Mishkin, Clark, et~al.]{radford2021learning}
Alec Radford, Jong~Wook Kim, Chris Hallacy, Aditya Ramesh, Gabriel Goh, Sandhini Agarwal, Girish Sastry, Amanda Askell, Pamela Mishkin, Jack Clark, et~al.
\newblock Learning transferable visual models from natural language supervision.
\newblock In \emph{International conference on machine learning}, pages 8748--8763. PMLR, 2021.

\bibitem[Ramesh et~al.(2021)Ramesh, Pavlov, Goh, Gray, Voss, Radford, Chen, and Sutskever]{ramesh2021zero}
Aditya Ramesh, Mikhail Pavlov, Gabriel Goh, Scott Gray, Chelsea Voss, Alec Radford, Mark Chen, and Ilya Sutskever.
\newblock Zero-shot text-to-image generation.
\newblock In \emph{International Conference on Machine Learning}, pages 8821--8831. PMLR, 2021.

\bibitem[Rombach et~al.(2021)Rombach, Blattmann, Lorenz, Esser, and Ommer]{rombach2021highresolution}
Robin Rombach, Andreas Blattmann, Dominik Lorenz, Patrick Esser, and Björn Ommer.
\newblock High-resolution image synthesis with latent diffusion models, 2021.

\bibitem[Ruiz et~al.(2023)Ruiz, Li, Jampani, Pritch, Rubinstein, and Aberman]{ruiz2023dreambooth}
Nataniel Ruiz, Yuanzhen Li, Varun Jampani, Yael Pritch, Michael Rubinstein, and Kfir Aberman.
\newblock Dreambooth: Fine tuning text-to-image diffusion models for subject-driven generation.
\newblock In \emph{Proceedings of the IEEE/CVF Conference on Computer Vision and Pattern Recognition}, pages 22500--22510, 2023.

\bibitem[Sajjadi et~al.(2018)Sajjadi, Bachem, Lucic, Bousquet, and Gelly]{sajjadi2018assessing}
Mehdi~SM Sajjadi, Olivier Bachem, Mario Lucic, Olivier Bousquet, and Sylvain Gelly.
\newblock Assessing generative models via precision and recall.
\newblock \emph{Advances in neural information processing systems}, 31, 2018.

\bibitem[Schroff et~al.(2015)Schroff, Kalenichenko, and Philbin]{schroff2015facenet}
Florian Schroff, Dmitry Kalenichenko, and James Philbin.
\newblock Facenet: A unified embedding for face recognition and clustering.
\newblock In \emph{Proceedings of the IEEE conference on computer vision and pattern recognition}, pages 815--823, 2015.

\bibitem[Seshadri et~al.(2023)Seshadri, Singh, and Elazar]{seshadri2023bias}
Preethi Seshadri, Sameer Singh, and Yanai Elazar.
\newblock The bias amplification paradox in text-to-image generation.
\newblock \emph{arXiv preprint arXiv:2308.00755}, 2023.

\bibitem[Smith et~al.(2023)Smith, Farinha, Hall, Kirk, Shtedritski, and Bain]{smith2023balancing}
Brandon Smith, Miguel Farinha, Siobhan~Mackenzie Hall, Hannah~Rose Kirk, Aleksandar Shtedritski, and Max Bain.
\newblock Balancing the picture: Debiasing vision-language datasets with synthetic contrast sets.
\newblock \emph{arXiv preprint arXiv:2305.15407}, 2023.

\bibitem[Struppek et~al.(2022)Struppek, Hintersdorf, and Kersting]{struppek2022biased}
Lukas Struppek, Dominik Hintersdorf, and Kristian Kersting.
\newblock The biased artist: Exploiting cultural biases via homoglyphs in text-guided image generation models.
\newblock \emph{arXiv preprint arXiv:2209.08891}, 2022.

\bibitem[Tanaka et~al.(2004)Tanaka, Kiefer, and Bukach]{TANAKA2004B1}
James~W Tanaka, Markus Kiefer, and Cindy~M Bukach.
\newblock A holistic account of the own-race effect in face recognition: evidence from a cross-cultural study.
\newblock \emph{Cognition}, 93\penalty0 (1):\penalty0 B1--B9, 2004.
\newblock ISSN 0010-0277.
\newblock \doi{https://doi.org/10.1016/j.cognition.2003.09.011}.
\newblock URL \url{https://www.sciencedirect.com/science/article/pii/S0010027703002336}.

\bibitem[Trabucco et~al.(2023)Trabucco, Doherty, Gurinas, and Salakhutdinov]{trabucco2023effective}
Brandon Trabucco, Kyle Doherty, Max Gurinas, and Ruslan Salakhutdinov.
\newblock Effective data augmentation with diffusion models.
\newblock In \emph{ICLR 2023 Workshop on Mathematical and Empirical Understanding of Foundation Models}, 2023.

\bibitem[Tsaban and Passos(2023)]{tsaban2023ledits}
Linoy Tsaban and Apolinário Passos.
\newblock Ledits: Real image editing with ddpm inversion and semantic guidance, 2023.

\bibitem[Tsao and Livingstone(2008)]{tsao2008mechanisms}
Doris~Y Tsao and Margaret~S Livingstone.
\newblock Mechanisms of face perception.
\newblock \emph{Annu. Rev. Neurosci.}, 31:\penalty0 411--437, 2008.

\bibitem[Zhang et~al.(2023)Zhang, Mo, Chen, Sun, and Su]{zhang2023MagicBrush}
Kai Zhang, Lingbo Mo, Wenhu Chen, Huan Sun, and Yu~Su.
\newblock Magicbrush: A manually annotated dataset for instruction-guided image editing, 2023.

\end{thebibliography}
\newpage
\appendix
\section{Proof of Proposition 1}
\setcounter{proposition}{0}

For the sake of completeness, we restate the proposition below

\begin{proposition}[Proposition]
Let $\delta \in (0,1)$. If 

\begin{equation}
\small
\abs{\trainingData} = \bigOh\bigg(\dimension^2\log(2) \log (2/\delta)\bigg[H^2(\sampleDist_a,\sampleDist_b)\bigg]^{-4}\bigg) 
\end{equation}
then we have 
\begin{equation}
\small
\Pr\bigg[\metric_{\totalVariation}(\hat{\sampleDist}_\trainingData,\sampleDist) > \frac{\abs{\alpha-\beta}}{2}H^2(\sampleDist_a,\sampleDist_b)\bigg] \geq 1  - \delta
\end{equation}
where $H^2$ is the squared Hellinger distance, and 

\begin{equation}
\small
\begin{split}
\small
 & H^2(\sampleDist_a,\sampleDist_b) = 
 \bigg(1-\frac{\abs{\vec{\Sigma}_a}^{\sfrac{1}{4}}\abs{\vec{\Sigma}_b}^{\sfrac{1}{4}} }{\abs{\frac{\vec{\Sigma}_a+\vec{\Sigma}_b}{2}}^{\sfrac{1}{2}}}\bigg)  \\
& \times \exp\bigg\{-\frac{1}{8}(\vec{\mu}_a-\vec{\mu}_b)^\transpose\bigg(\frac{\vec{\Sigma}_a+\vec{\Sigma}_b}{2}\bigg)^{-1}(\vec{\mu}_a-\vec{\mu}_b)\bigg\}
\end{split}
\end{equation} 

if $\sampleDist_a,\sampleDist_b$ are each multivariate Gaussians.
\end{proposition}

Let us now prove the proposition:

\begin{proof}

We will leverage triangle inequality to lower bound $\metric_{\totalVariation}(\hat{\sampleDist}_\trainingData,\sampleDist)$. That decomposition is 

\begin{equation}\label{eq:WassersteinTriangle}
    \metric_{\totalVariation}(\sampleDist_\trainingData,\sampleDist) \leq \metric_{\totalVariation}(\sampleDist_\trainingData,\hat{\sampleDist}_\trainingData) + \metric_{\totalVariation}(\hat{\sampleDist}_\trainingData,\sampleDist)
\end{equation}

The squared Hellinger distance lower bounds discrepancy. Hence, we have

\begin{align}
    \metric_{\totalVariation}(\sampleDist_\trainingData,\sampleDist) &= \sup_{\placeholder \in \mathcal{F}}~\abs{\sampleDist_\trainingData(\placeholder)-\sampleDist(\placeholder)}\\
    \begin{split}
         &= \sup_{\placeholder \in \mathcal{F}}~\Big\lvert\beta\sampleDist_a(\placeholder)+(1-\beta)\sampleDist_b(\placeholder) \\
        &\qquad -(\alpha\sampleDist_a(\placeholder)+(1-\alpha)\sampleDist_b(\placeholder))\Big\rvert
    \end{split}\\
    &=  \abs{\alpha-\beta}\metric_{\totalVariation}(\sampleDist_a,\sampleDist_b) \\
    &\geq \abs{\alpha-\beta}H^2(\sampleDist_a,\sampleDist_b)
\end{align}

We also need an upper bound on $\metric_{\totalVariation}(\sampleDist_\trainingData,\hat{\sampleDist}_\trainingData)$. That upper-bound is sourced from Ashtiani et al.~\cite{ashtiani2018sample}.

\begin{theorem}[PAC learnability of Gaussian Mixtures~\cite{ashtiani2018sample}]
    Let $\epsilon,\delta > 0$. Let $\trainingData$ be drawn i.i.d. from $\placeholderDist$, and let $\hat{\placeholderDist}$ be the distribution learned from $\trainingData$. With probability at least $1-\delta$, given $\bigOh\big(k\dimension^2\log(k) \log (k/\delta)/(\epsilon)^4\big)$ samples in $\trainingData$., a realizable mixture of $k$-Gaussian mixtures is $\epsilon$-learned, i.e
    \begin{equation}
    \Pr[\metric_{\totalVariation}(\hat{\placeholderDist},\placeholderDist)\leq \epsilon] \geq 1-\delta 
    \end{equation}
\end{theorem}

Thus, combining the theorem and triangle inequality, we arrive at a high probability lower-bound on $\metric_{\totalVariation}(\hat{\sampleDist}_\trainingData,\sampleDist)$. When $\abs{\trainingData} = \bigOh\big(\dimension^2\log(2) \log (2/\delta)/\epsilon^4\big)$, then

\begin{equation}
\Pr\bigg[\metric_{\totalVariation}(\hat{\sampleDist}_\trainingData,\sampleDist) > \abs{\alpha-\beta}H^2(\sampleDist_a,\sampleDist_b) - \epsilon\bigg] \geq 1  - \delta
\end{equation}

Setting $\epsilon = \frac{\abs{\alpha-\beta}}{2}H^2(\sampleDist_a,\sampleDist_b)$, we arrive at the proposition.
 \end{proof}

\section{Statistical Tests}
We evaluate several null hypotheses in our evaluation. Further details, including $p$-values, associated with our statistical tests are provided below. 
\subsection{Null Hypotheses 1-3}
In our evaluation, we discuss the use of one-way ANOVA to reject \cref{nullhyp:sourcequality,nullhyp:targetquality,nullhyp:sourcetargetdrop}.  Table~\ref{tab:nullhypothesis123pvalues} presents the corresponding $p$-values for Realism Celebrities, Realism non-Celebrities, and SDv2.1 Celebrities datasets.

\begin{table}[t]
\centering
\begin{tabular}{lccc}
\toprule
\textbf{Dataset} & \textbf{\Cref{nullhyp:sourcequality}} & \textbf{\Cref{nullhyp:targetquality}} & \textbf{\Cref{nullhyp:sourcetargetdrop}}\\
\midrule

SDv2.1 Celebrities 
 & $0.498$ & $0.573$ & $0.488$ \\
\midrule
Realism Celebrities 
&$0.000474$ & $6.31 \times 10^{-25}$ & $4.02 \times 10^{-28}$\\
\midrule
Realism Non-Celebrities 
& $0.0306$ & $5.16 \times 10^{-5}$ & $2.47 \times 10^{-5}$ \\
\bottomrule
\end{tabular}
\caption{$p$-values associated with one-way ANOVAs on \cref{nullhyp:sourcequality,nullhyp:targetquality,nullhyp:sourcetargetdrop} for  Realism Celebrities, Realism non-Celebrities, and SDv2.1 Celebrities datasets }\label{tab:nullhypothesis123pvalues}

\end{table}

\subsection{Null Hypotheses 4 and 5}

In \cref{tab:spearmanmetrics}, we stated Spearman correlation coefficients between quantitative image metrics and user surveys. The corresponding $p$-values are stated in \cref{tab:Spearmanpvalues}.

\begin{table}[t]
\centering

\begin{tabular}{lccc}
\toprule
\multirow{2}{*}{\textbf{Dataset}} & \multicolumn{2}{c}{\textbf{\Cref{nullhyp:imagesimilariymetricCorrelation}}} & \textbf{\Cref{nullhyp:directionalsimilaritymetricCorrelation}}\\
\cmidrule{2-4}
& \textbf{CLIP-I} & \textbf{DINO-I} & \textbf{CLIP Directional}\\
\midrule

SDv2.1 Celebrities 
 & $6.264 \times 10^{-18}$ & $7.35 \times 10^{-14}$ & $3.55 \times 10^{-10}$ \\
\midrule
Realism Celebrities 
&$5.74 \times 10^{-32}$ & $1.022 \times 10^{-100}$ & $3.27 \times 10^{-13}$\\
\midrule
Realism Non-Celebrities 
& $3.26 \times 10^{-60}$ & $2.18 \times 10^{-9}$ & $6.64 \times 10^{-21}$ \\
\bottomrule
\end{tabular}
\caption{$p$-values associated with \cref{nullhyp:imagesimilariymetricCorrelation,nullhyp:directionalsimilaritymetricCorrelation}.}\label{tab:Spearmanpvalues}

\end{table}

\section{Data Generation Pipeline}
We provide further details of our data generation pipeline, include more examples of the different datasets produced across all demographics, and discuss the approach's limitations.
\subsection{Generating Multiple Source Image Variations}
\begin{figure}[h]
    \centering
\includegraphics[height=5.5in]{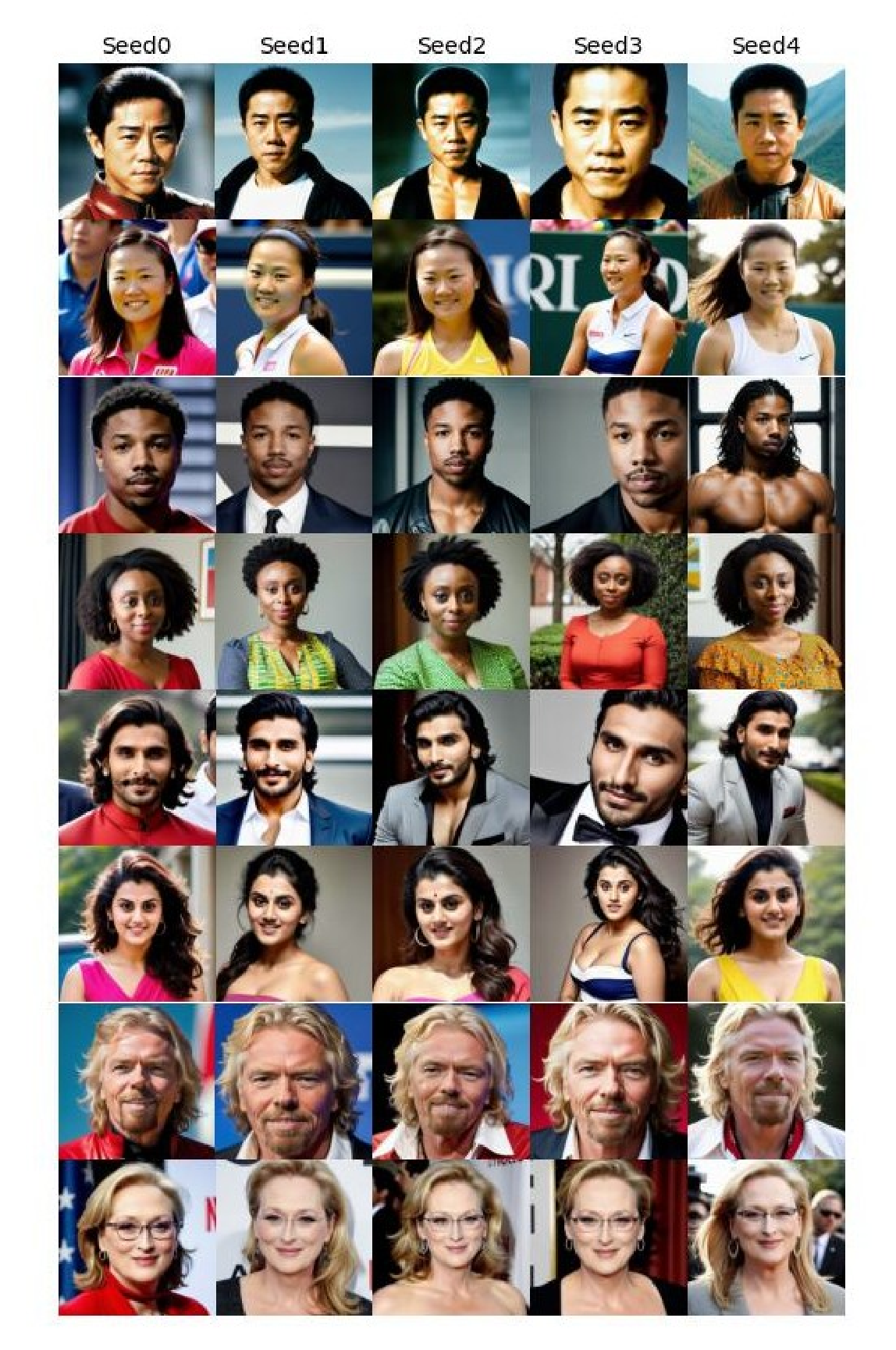}
    \caption{Variations of Source Faces of Realism Celebrities across all demographics}
    \label{fig:CelebSourceRealism}
\end{figure}

\begin{figure}[h]
    \centering
    \includegraphics[height=5.5in]{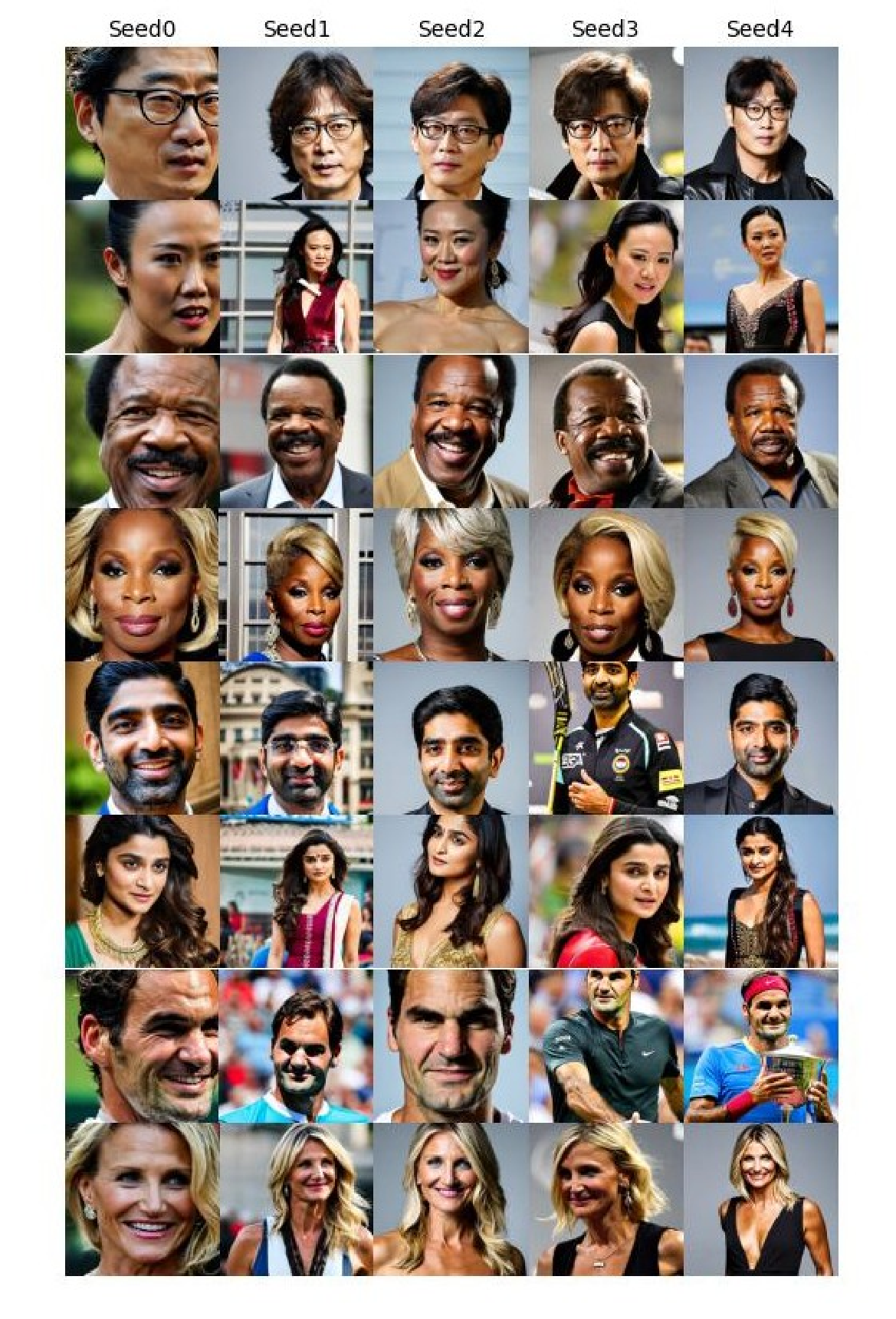}
    \caption{Variations of Source Faces of SDv2.1 Celebrities across all demographics}
    \label{fig:CelebSourceSDv2.1}
\end{figure}

\begin{figure}[h]
    \centering
    \includegraphics[height=5.5in]{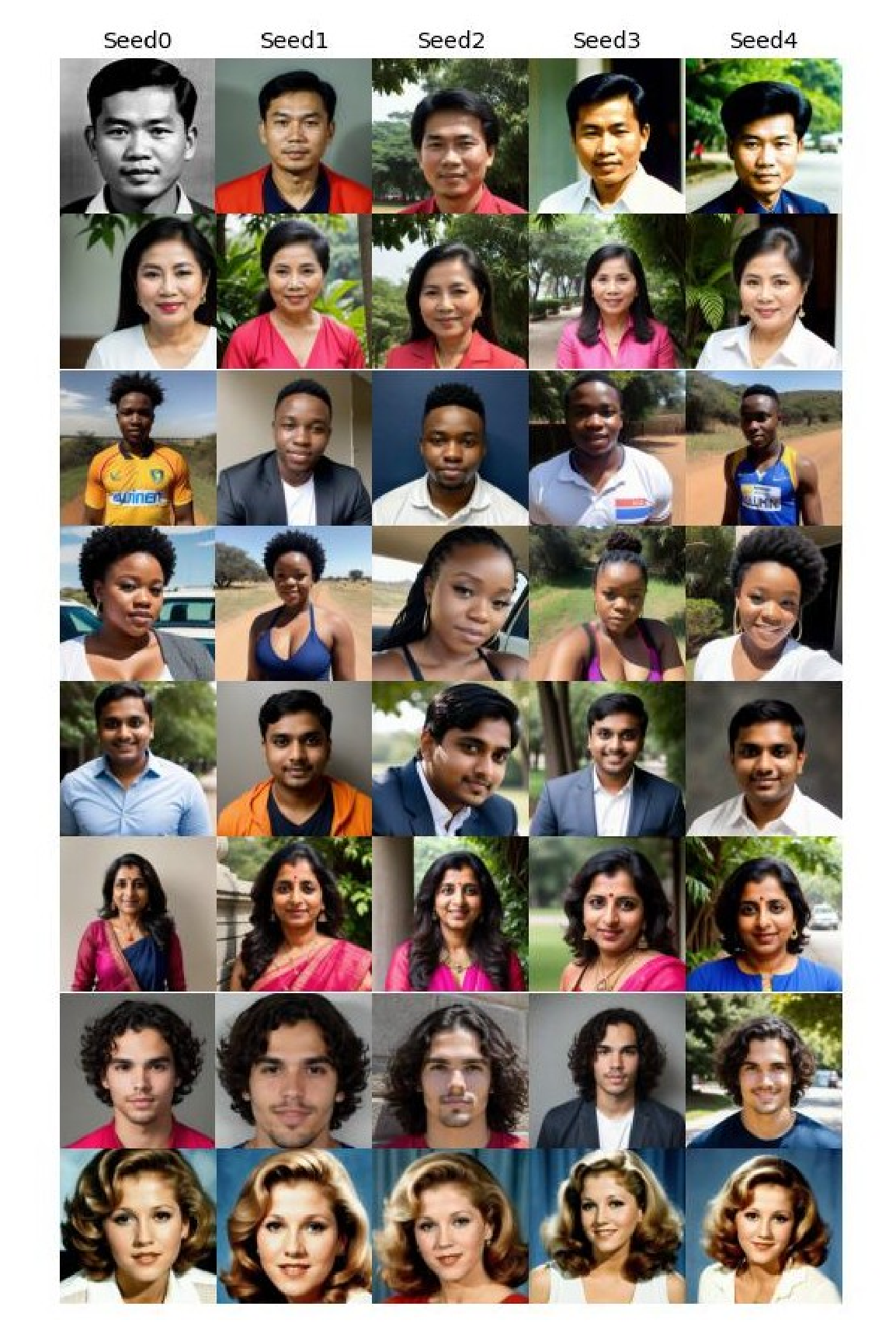}
    \caption{Variations of Source Faces of Realism Non-Celebrities across all demographics}
    \label{fig:NonCelebSourceRealism}
\end{figure}

We generate 5 source images~(variations) for each identity for celebrities using Realism and SDv2.1, and non-celebrities using Realism. We achieve this by keeping a fixed prompt and varying the seed. A sample of the variations obtained for identities across all the eight demographics for Realism-Celebrities, SDv2.1 Celebrities, and Realism Non-Celebrities datasets can be seen in \cref{fig:CelebSourceRealism,fig:CelebSourceSDv2.1,fig:NonCelebSourceRealism}. This approach increased the dataset size and diversity of different identities across all demographics. However, it's worth noting that there were cases where both Realism and SDv2.1 failed to adhere to the prompt and retain identities for different seeds. 

\begin{figure}[h]
    \centering
    \begin{subfigure}[b]{\textwidth}
        \centering \includegraphics[ height=2.5in]{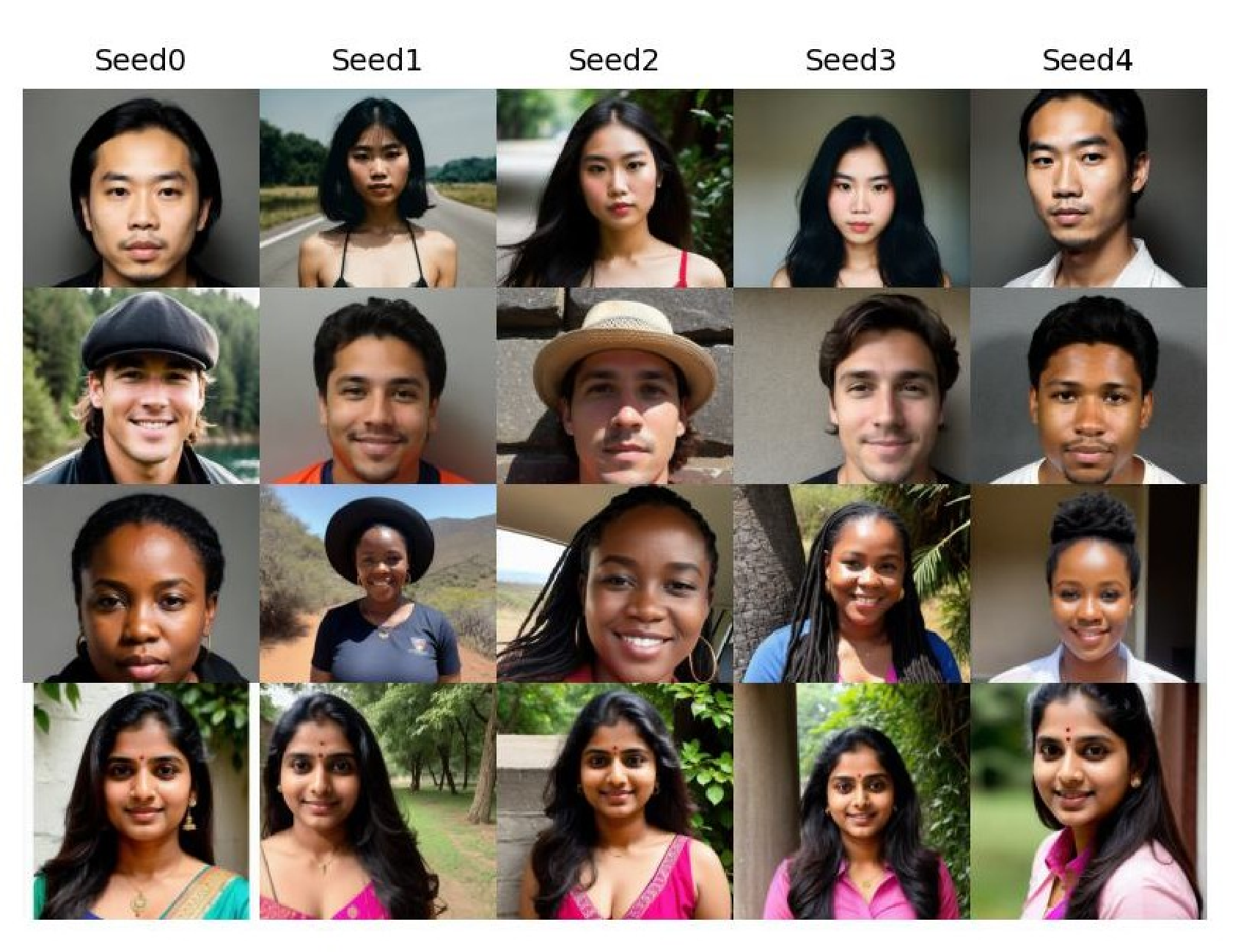}
        \caption{Realism Non-Celebrities' partial identity mismatch across variations}
        \label{fig:Identitymismatch}
    \end{subfigure}
    \begin{subfigure}[b]{\textwidth}
    \centering
    \includegraphics[ height=1.5in]{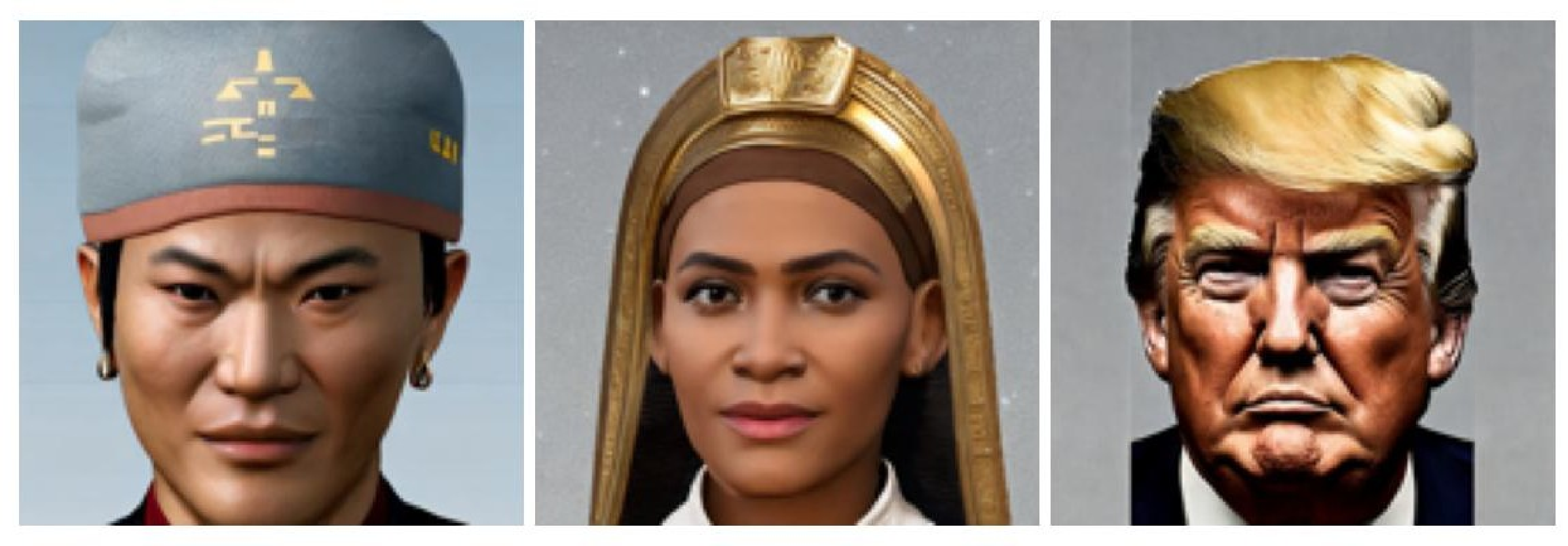}
        \caption{Less-realistic faces obtained in SDv2.1 Celebrities data}
        \label{fig:lessrealistic}
    \end{subfigure}
    \begin{subfigure}[b]{\textwidth}
        \centering
        \includegraphics[height=1.8in]{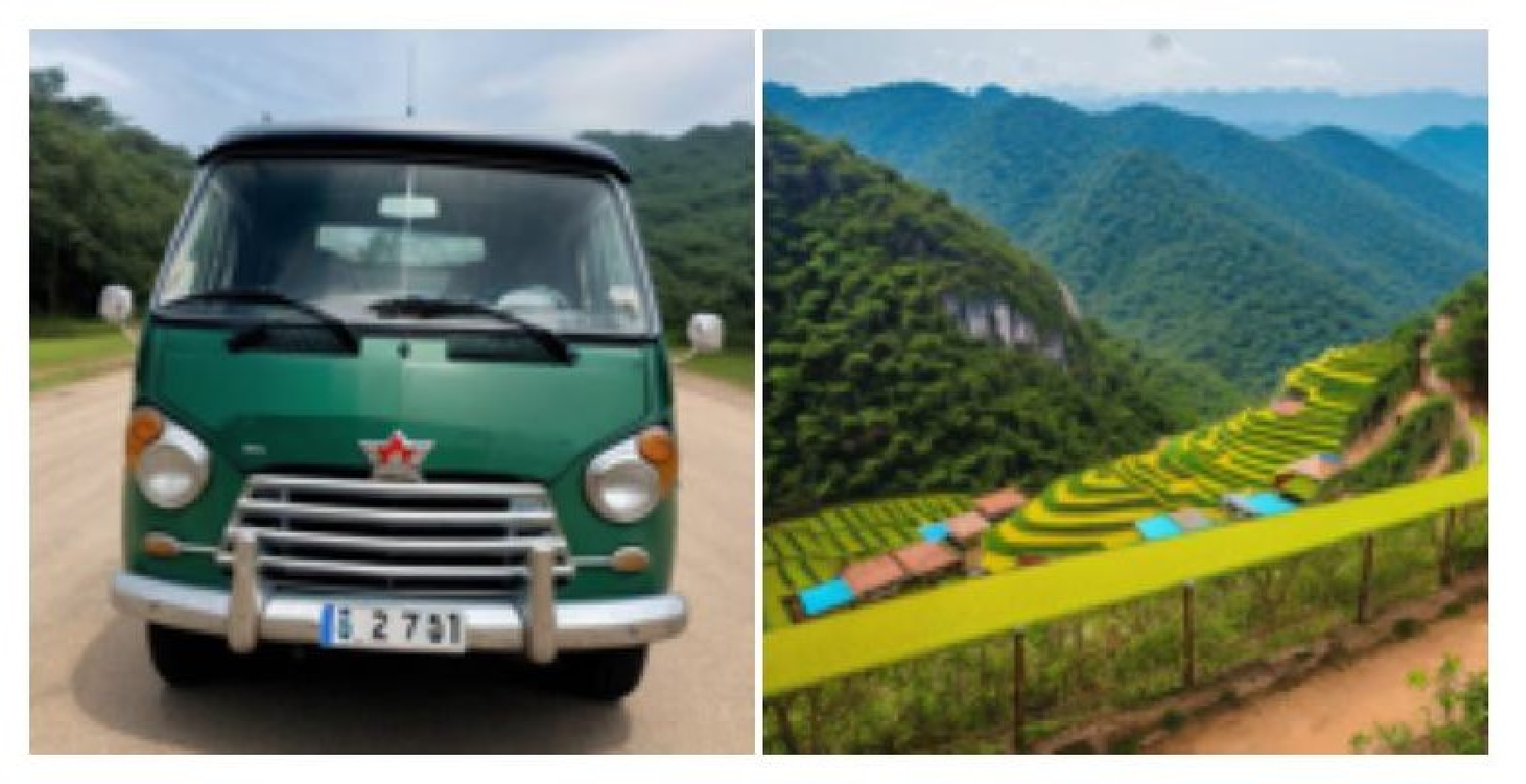}
        \caption{Non-faces obtained in a variation of two identities in Realism Non-Celebrities data}
        \label{fig:non-faces}
    \end{subfigure}
    \begin{subfigure}[b]{\textwidth}
        \centering
        \includegraphics[height=2in]{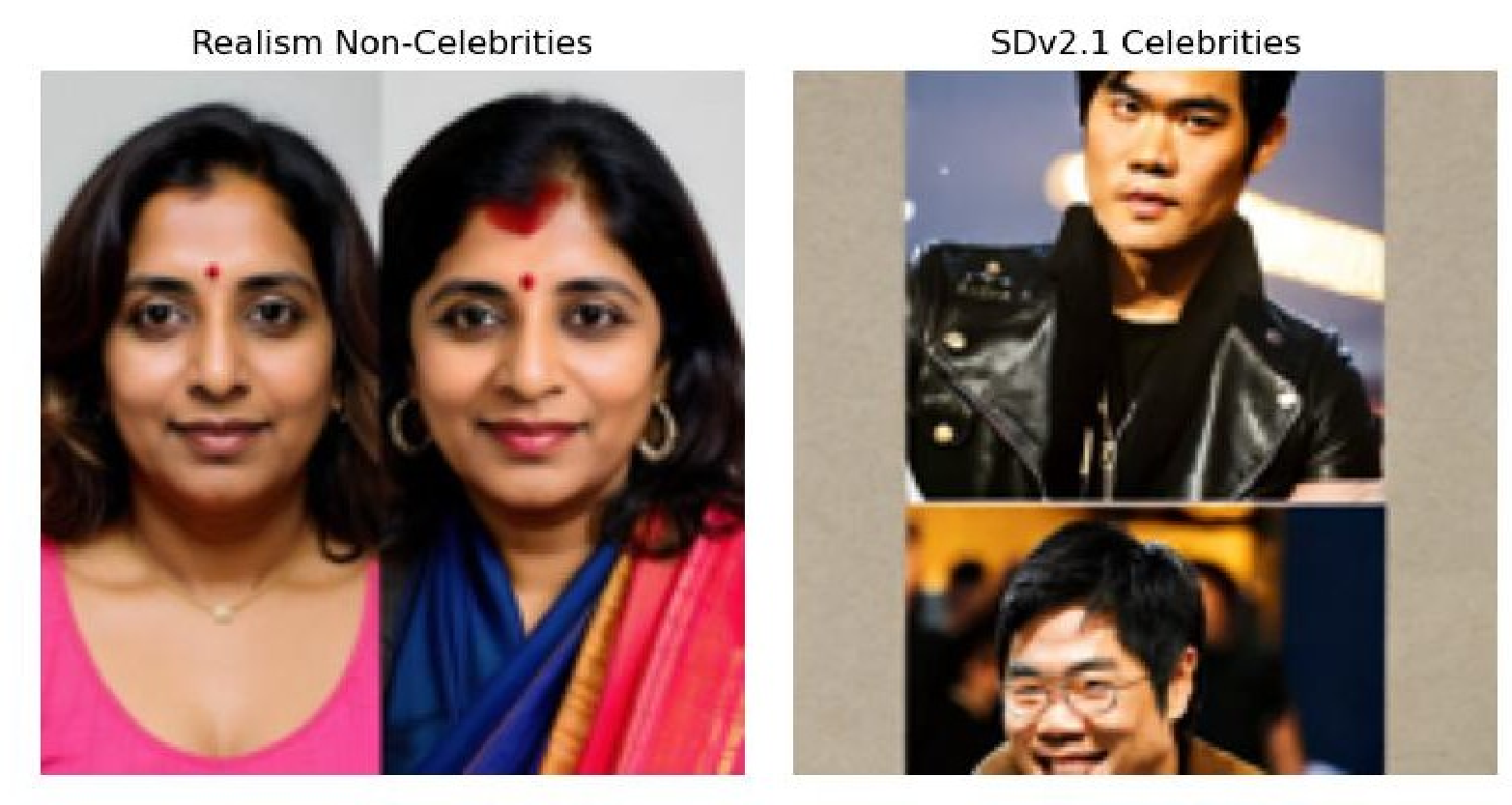}
        \caption{Multiple faces synthesized for a single seed and prompt in Realism Non-Celebrities and SDv2.1 Celebrities data}
        \label{fig:multiplefaces}
    \end{subfigure}
    \caption{Limitations of generating multiple source variations}
    \label{fig:combined}
\end{figure}

\begin{figure}[h]
    \centering
    \includegraphics[height=2.5in]{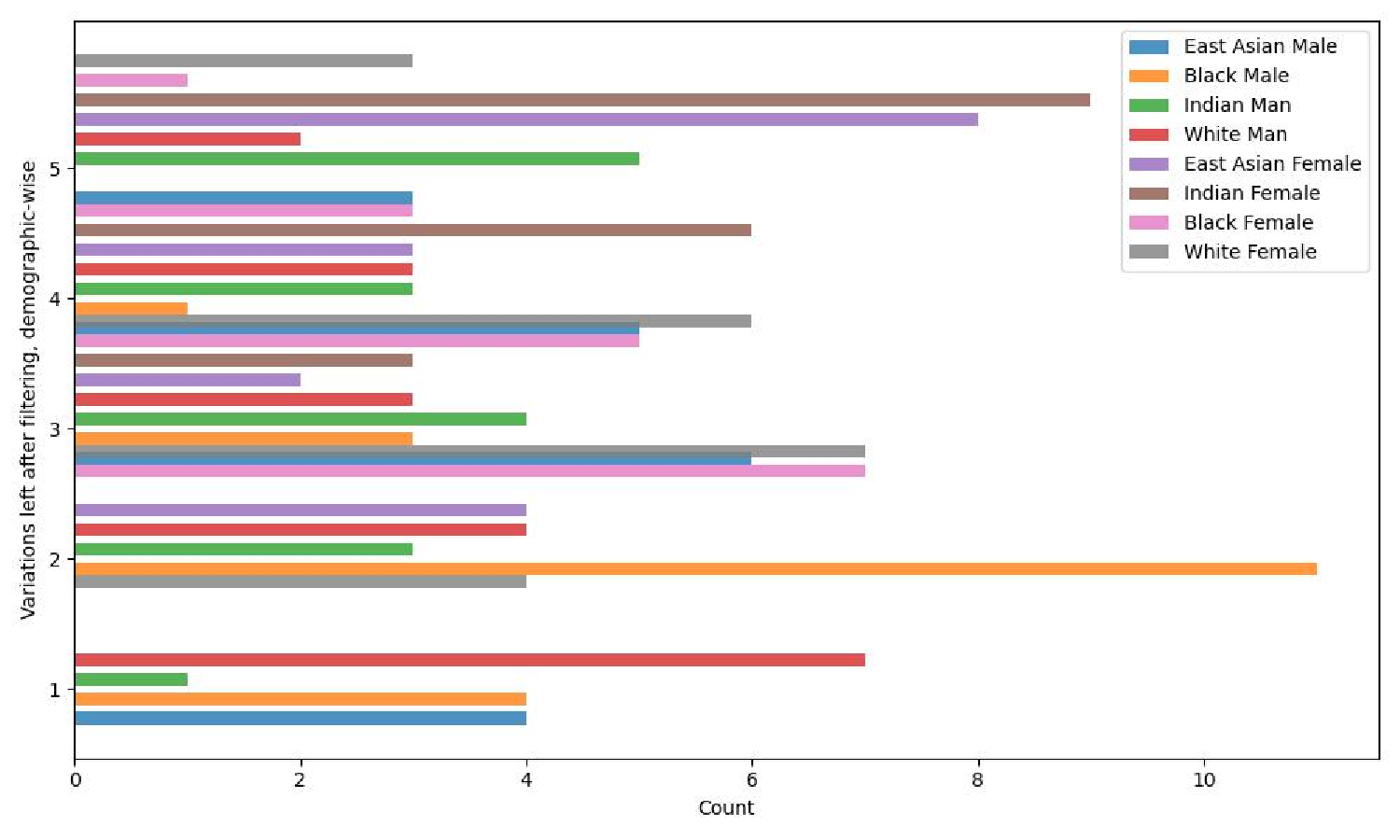}
    \caption{Demographic-wise distribution of variations left per identity after manual filtering of Realism Non-celebrities}
    \label{fig:filtering}
\end{figure}

We observed the following issues in our data generation pipeline: 
\begin{itemize}
    \item \textbf{Non-celebrity variations required manual filtering} We observed that Realism and SDv2.1 struggle to consistently synthesize non-celebrity faces of the same identity for the same prompt across all 5 seeds. A few examples of this in Realism Non-Celebrities can be seen in \cref{fig:Identitymismatch}. This forced us to drop the non-celerity data generated using SDv2.1 altogether and manually filter the variations of Realism Non-celebrities before the evaluation steps. The filtering process involved two steps. In the first step, identities that exhibited excessive facial similarity with other identities were dropped. Next, variations of an identity that didn't match the other variations were discarded. In some cases, we also noticed variations with a different sex than what was expected. 
    
    For the final dataset, we only included the transformed faces of the filtered variations for each identity. After filtering, we had 143 out of 160 initial identities, comprising 450 source faces. The distribution of the number of variations per identity left after manual filtering can be viewed in \cref{fig:filtering}. The demographics `Asian Female' and `Indian Female' retained 4 or 5 variations for most identities. On the other hand, the demographics `White Male' and `Black Male` mostly retain only 1 to 2 variations after filtering.

    This problem was also present in Realism Celebrities and SDv2.1 celebrities but to a much lesser extent. So, the generated data was directly for evaluation.
    
    \item \textbf{Less realistic faces} This effect was mostly observed for SDv2.1 celebrities. Many faces were less realistic and looked cartoonish despite emphasizing the synthesis of a realistic face in the prompt and negative prompt. We suspect this could be due to the nature of the training data of SDv2.1, which consists of data sourced from the internet, including non-realistic faces. \cref{fig:lessrealistic} contains examples of this observation.
    \item \textbf{Non-faces instead of faces} We also noticed a handful of cases across the three datasets, where non-faces could be found instead of human faces. As shown in \cref{fig:non-faces}, this can include objects like a van and a scenery instead of a human face.
    \item \textbf{Multiple Face Generation} Finally, we detected a handful of cases for Realism Non-Celebrities and SDv2.1 celebrities, where multiple faces were synthesized for a variation of an identity~(\cref{fig:multiplefaces}). 
    
\end{itemize}

\subsection{Generating transformed images with SEGA}

\begin{figure}[h]
    \centering
    \includegraphics[height=5.5in]{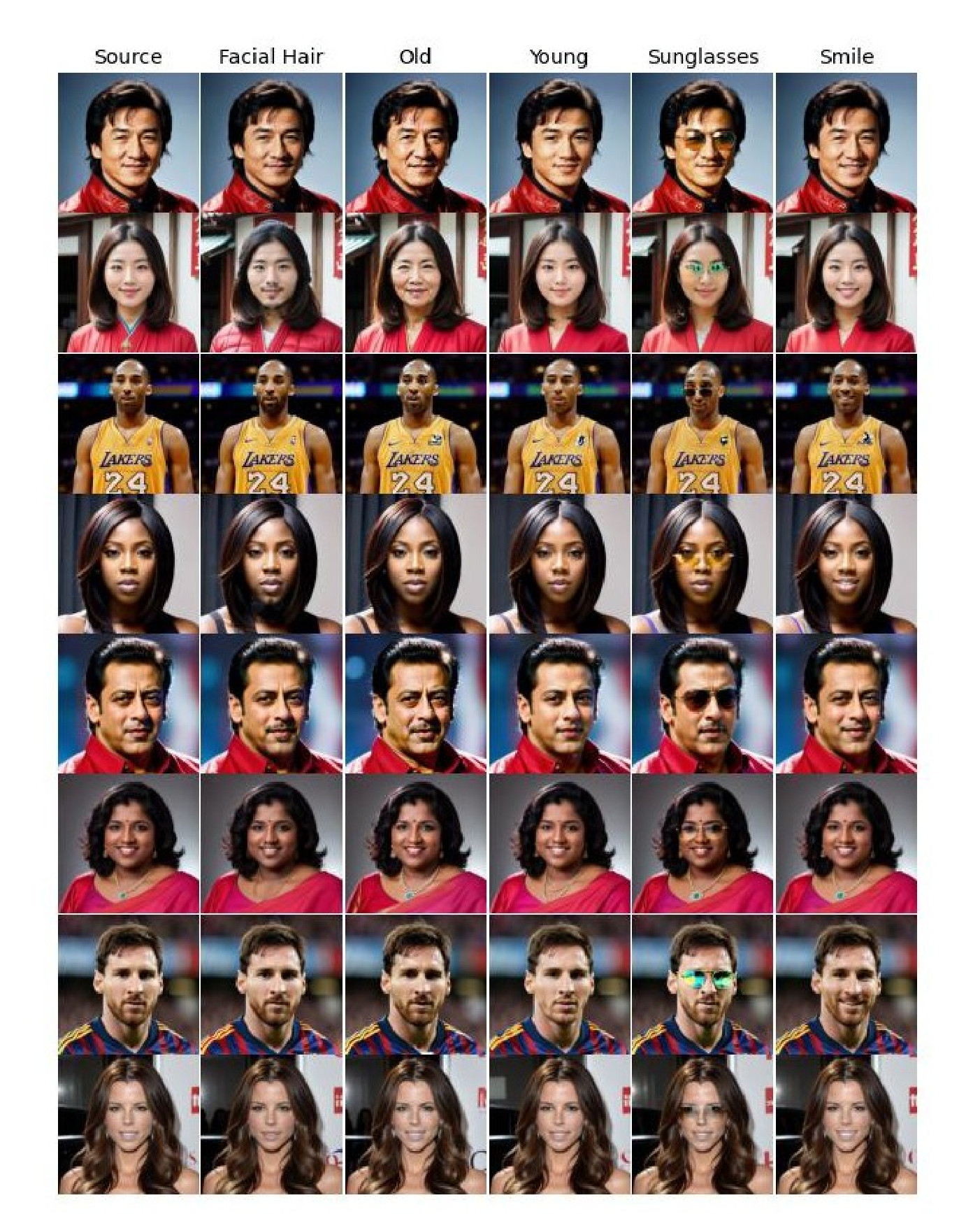}
    \caption{Transformed Faces for Realism Celebrities across all demographics. The first column indicates the source faces, and the rest indicate the transformed faces after applying the attributes 'Facial Hair', 'Old', 'Young', 'Sunglasses', and 'Smile' using SEGA}
    \label{fig:CelebGridRealism}
\end{figure}

\begin{figure}[h]
    \centering   \includegraphics[height=5.5in]{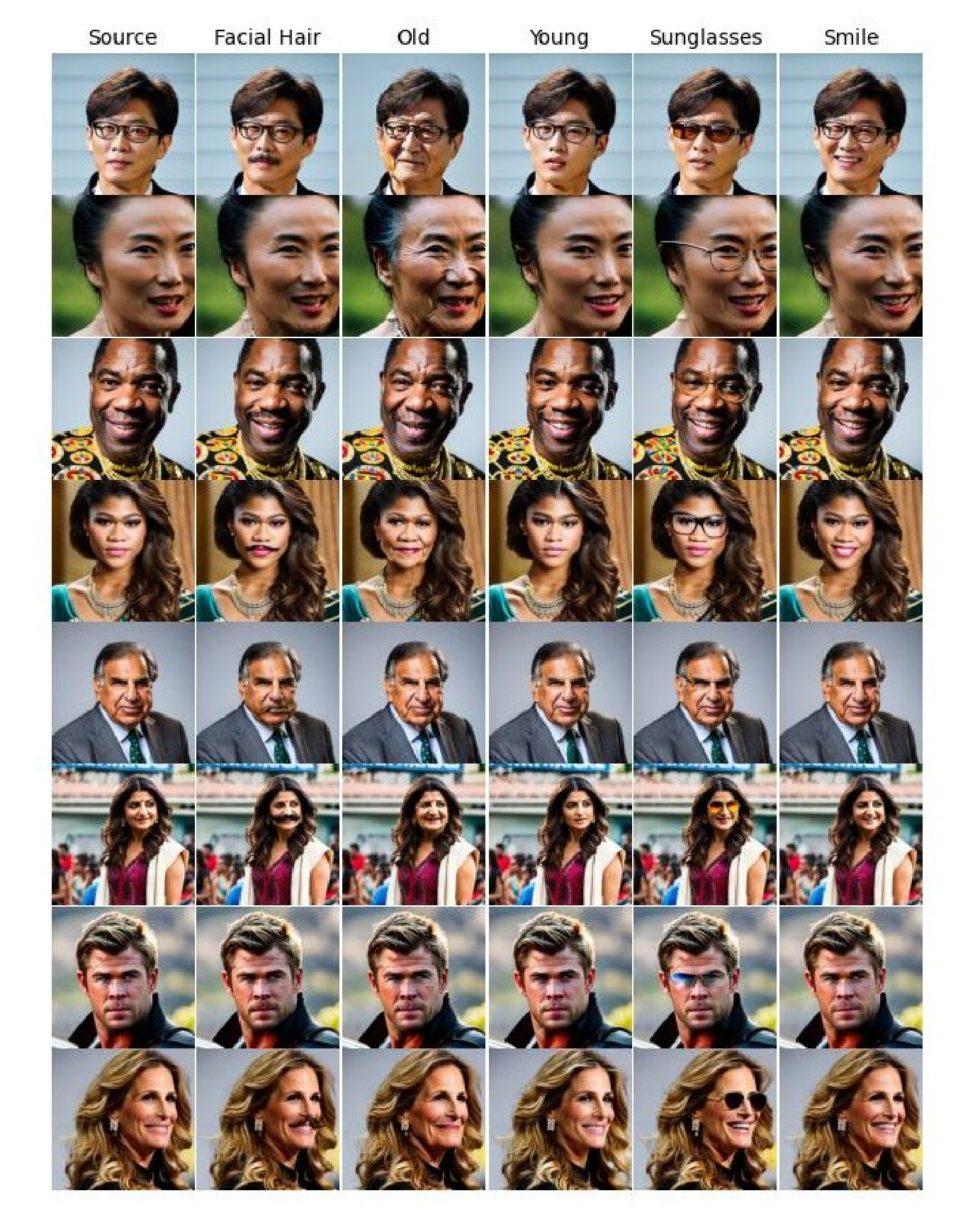}
    \caption{Transformed Faces for SDv2.1 Celebrities across all demographics. The first column indicates the source faces, and the rest indicate the transformed faces after applying the attributes 'Facial Hair', 'Old', 'Young', 'Sunglasses', and 'Smile' using SEGA}
    \label{fig:CelebGridSDv2.1}
\end{figure}

\begin{figure}[h]
    \centering
    \includegraphics[height=5.5in]{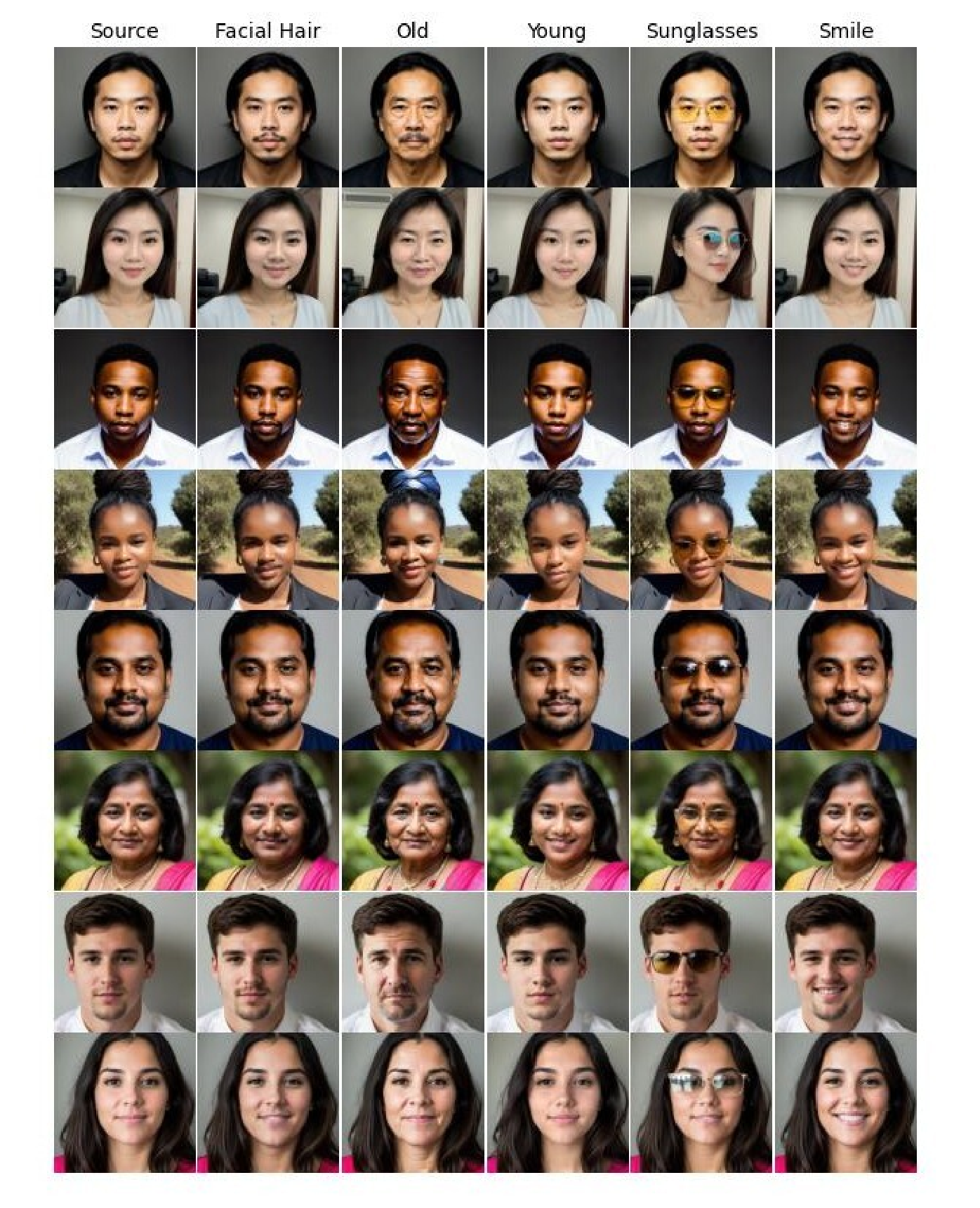}
    \caption{Transformed Faces for Realism Non-Celebrities across all demographics. The first column indicates the source faces, and the rest indicate the transformed faces after applying the attributes 'Facial Hair', 'Old', 'Young', 'Sunglasses', and 'Smile' using SEGA}
    \label{fig:NonCelebGridRealism}
\end{figure}

We generate transformed faces using SEGA with Realism and SDv2.1 for the semantic concepts~(attributes) `old', `young', `facial hair', `sunglasses', and `smile'. Although `old' and `young' are social and biological constructs of a human face, we use them along with the other attributes. This is especially useful in testing the performance of a face verification system for a face and its older and younger versions. Examples of the transformed faces for Realism Celebrities, SDv2.1 Celebrities, and Realism Non-Celebrities can be seen in \cref{fig:CelebGridRealism,fig:CelebGridSDv2.1,fig:NonCelebGridRealism}. 

The main bottleneck in obtaining transformed faces with SEGA was getting a set of SEGA hyperparameters to work for all the identities across all demographics for Realism and SDv2.1. As this task is strenuous, we chose a set of hyperparameters for each attribute that can work well across identities and models. We tuned the `edit guidance scale', `edit threshold', and `edit weights' with the help of a small set of held-out identities. The values of these hyperparameters can be found in \cref{table:SEGAhyperparams}. The other hyperparameters, `edit warmup-steps', `edit momentum-scale', and `edit momentum-beta', are set to $10$, $0.3$, and $0.6$, respectively. For the attributes `old' and `young', a negative concept is added to ensure the attribute is applied correctly. In SEGA, a negative concept has a reverse editing effect. We use $50$ diffusion steps and a guidance scale of $7.5$ for Realism and SDv2.1.  

\begin{figure}[h]
    \centering
    \includegraphics[width=\textwidth]{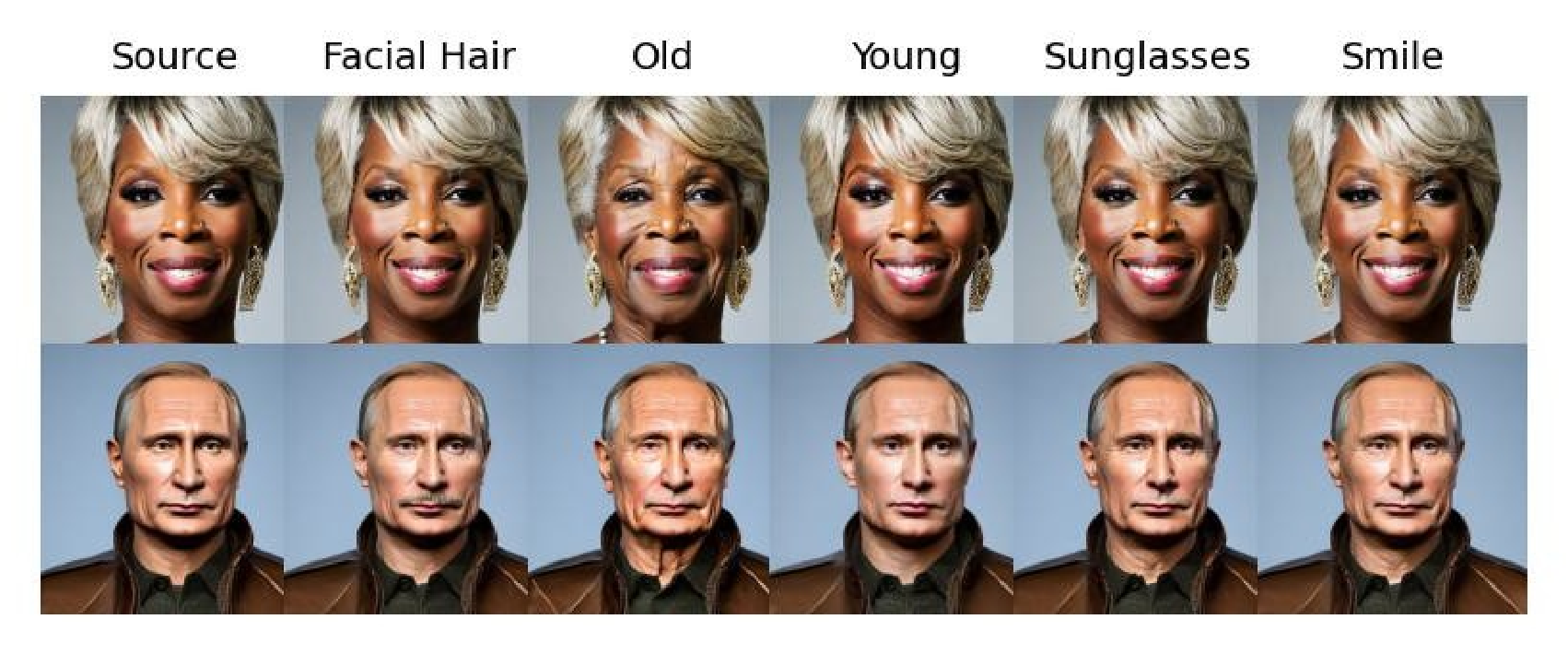}
    \caption{Transformed faces that don't contain the requested attribute. We can see in the first row that there is no change for the attributes `facial hair' and `sunglasses', and no change for the attributes `sunglasses' and 'smile' in the second row.}
    \label{fig:nochange}
\end{figure}

\begin{figure}[h]
    \centering
    \includegraphics[height=2.5in]{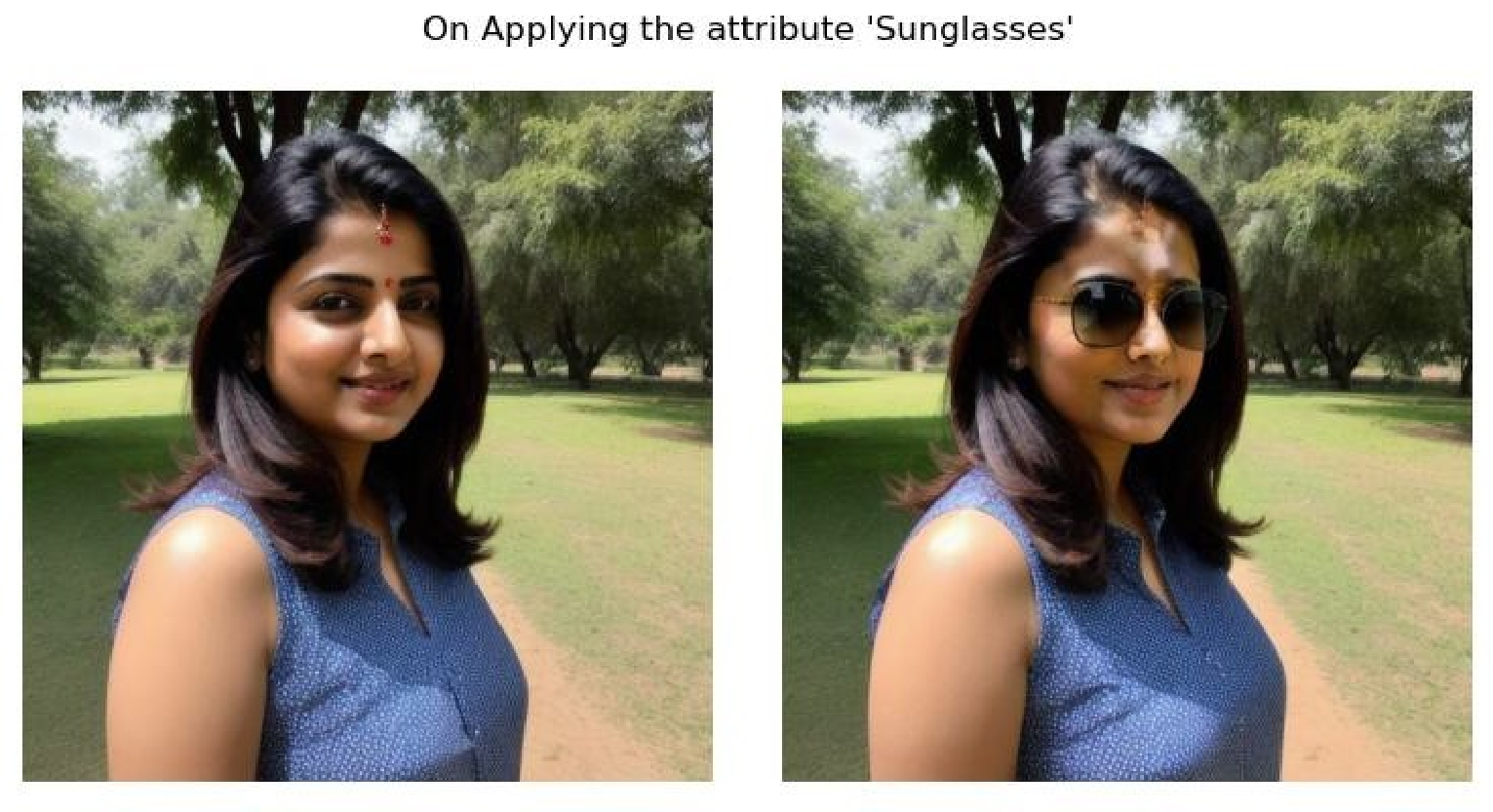}
    \caption{Face quality deterioration of transformed faces: Here, the transformed face is unable to retain the features on the forehead after adding the attribute `sunglasses'}
    \label{fig:Deteriorate}
\end{figure}

\cref{fig:nochange} shows examples of the attribute `facial hair' and 'sunglasses' in the SDv2.1 Celebrity dataset, where the transformed faces don't contain the attribute requested using SEGA. This is the most common drawback we observe across datasets. There are a handful of transformed faces that deteriorate on applying an attribute with SEGA, and an example can be seen in \cref{fig:Deteriorate}. 

\begin{table}[]
\centering

\begin{tabular}{ccccc}
\toprule
\textbf{Attribute} & \textbf{\begin{tabular}[c]{@{}c@{}}Edit \\ caption\end{tabular}} & \textbf{\begin{tabular}[c]{@{}c@{}}Edit Guidance \\ Scale\end{tabular}} & \textbf{\begin{tabular}[c]{@{}c@{}}Edit \\ Threshold\end{tabular}} & \textbf{\begin{tabular}[c]{@{}c@{}}Edit\\  Weights\end{tabular}} \\ \hline
Old & \begin{tabular}[c]{@{}c@{}}{[}`old face',\\ `young face'{]}\end{tabular} & {[}10, 8{]} & {[}0.96, 0.96{]} & {[}1, 0.6{]} \\ \hline
Young & \begin{tabular}[c]{@{}c@{}}{[}`young face',\\ `old face'{]}\end{tabular} & {[}10, 8{]} & {[}0.96, 0.96{]} & {[}1, 0.6{]} \\ \hline
Sunglasses & \begin{tabular}[c]{@{}c@{}}`sunglasses, \\ wearing sunglasses'\end{tabular} & 9 & 0.99 & 1 \\ \hline
Smile & \begin{tabular}[c]{@{}c@{}}`smile, \\ smiling'\end{tabular} & 8 & 0.99 & 0.8 \\ \hline
Facial Hair & \begin{tabular}[c]{@{}c@{}}`facial hair,\\  beard'\end{tabular} & 5 & 0.98 & 1 \\ \bottomrule
\end{tabular}%

\caption{Hyperparameters for SEGA. Square brackets indicate the usage of an additional negative semantic concept}
\label{table:SEGAhyperparams}
\end{table}

\section{User Survey}

\begin{figure*}[t]
    \centering
    \begin{subfigure}[t]{0.5\textwidth}
        \centering
        \includegraphics[width=\textwidth]{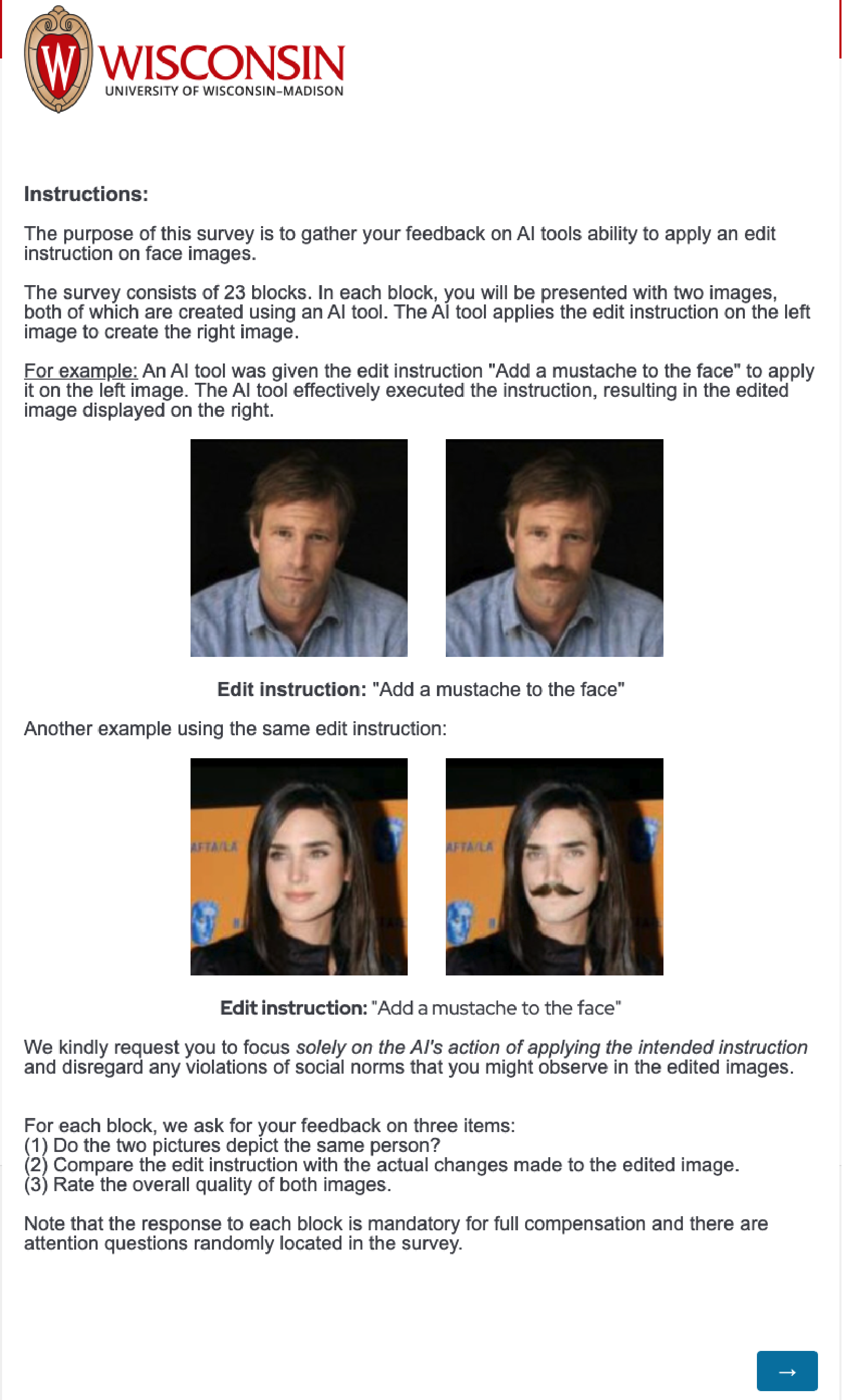}
        \caption{Survey Instructions}
        \label{fig:survey_instruction}
    \end{subfigure}
    \hfill
    \scalebox{.9}{
    \begin{subfigure}[t]{0.45\textwidth}
        \centering
        \includegraphics[width=\textwidth, height=6in]{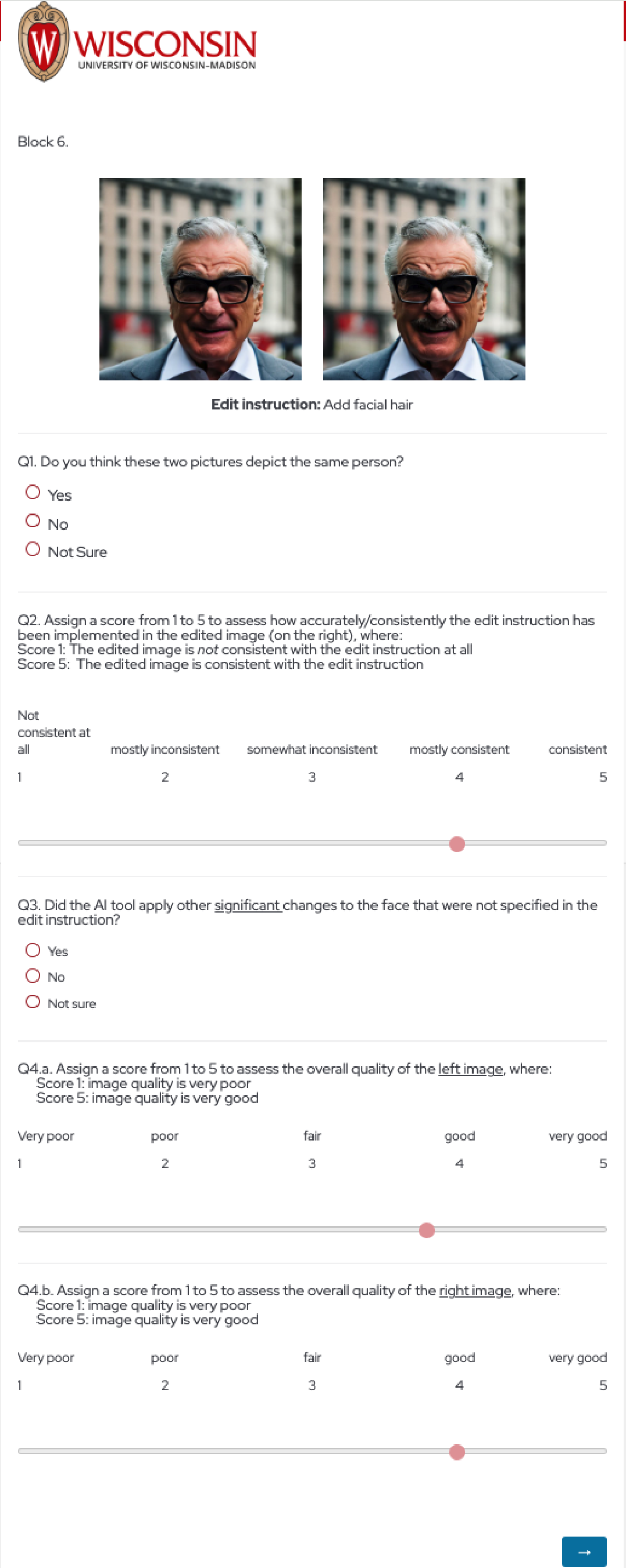}
        \caption{An example of a survey block}
        \label{fig:survey_q}
    \end{subfigure}}
    \caption{User survey instructions and example block of questions. }
    \label{fig:survey_details}
\end{figure*}

Figure~\ref{fig:survey_details} shows more details about the survey design. The survey instructions (\ref{fig:survey_instruction}) encourage participants to focus solely on the correctness of the edits and to disregard any violations of social norms they may observe in the images. This instruction is important since our work's ultimate goal is to generate a dataset that facilitates an assessment of face recognition applications in both in-distribution and out-of-distribution scenarios. Out-of-distribution images may possess attributes that seem unusable to participants.  

Each participant answered 21 blocks of questions, an example block is shown in \cref{fig:survey_q}, and at least one additional block of attention check questions. The placement of the attention check block was randomized within the survey flow. In each survey block, participants viewed two generated images and were asked to (Q1) verify the identity retention, (Q2) rate transformation correctness, (Q3) point out if the generative model applied \textit{unspecified} changes to the image, and (Q4) assess the overall quality of the images. Although we have not yet utilized the information collected from Q3, we consider it valuable for future research exploring spurious correlations in generative models.

Table~\ref{tab:demographic_distribution} shows the distribution of survey participants' demographics in terms of age, race, sex, and education.

\begin{table}[ht]
\centering
\caption{Demographic distribution of the three survey participants. D1: Realism Non-Celebrities, D2: Realism Celebrities, D3: SDv2.1 Celebrities.}
\label{tab:demographic_distribution}

\subcaption*{Age Distribution}
\begin{tabular}{@{}lccccc@{}}
\toprule
Survey    & 18 - 24 & 25 - 34 & 35 - 44 & 45 - 54 & 55 - 64 \\ \midrule
D1  & 33.0\%  & 43.7\%  & 12.6\%  & 7.8\%   & 2.9\%   \\
D2  & 37.0\%  & 47.5\%  & 9.6\%   & 6.0\%   & --      \\
D3  & 38.3\%  & 42.0\%  & 12.3\%  & 6.2\%   & 1.2\%   \\ \bottomrule
\end{tabular}

\bigskip 

\subcaption*{Race Distribution}
\begin{tabular}{@{}lcccccc@{}}
\toprule
Survey    & White & Black & Hispanic & Other & {East Asian} & {South Asian} \\ 
\midrule
D1 & 63.1\% & 27.2\% & 4.9\%    & 3.9\% & 1.0\%     & -- \\
D2 & 65.7\% & 14.3\% & 14.1\%   & 3.6\% & 1.2\%      & 1.2\%       \\
D3  & 53.1\% & 23.5\% & 13.6\%   & 3.7\% & 1.2\%      & 2.5\%       \\ 
\bottomrule
\end{tabular}

\bigskip 

\subcaption*{Sex Distribution}
\begin{tabular}{@{}lcccc@{}}
\toprule
Survey    & Male & Female & Other & N/A \\ \midrule
D1  & 51.5\% & 47.6\% & 1.0\% & --               \\
D2  & 50.0\% & 47.6\% & 1.2\% & 1.2\%            \\
D3  & 48.1\% & 49.4\% & 2.5\% & --               \\ \bottomrule
\end{tabular}

\bigskip

\subcaption*{Education Distribution}
\begin{tabular}{@{}lcccccc@{}}
\toprule
Survey & Less than high school & High school & Some college & 2 year degree & 4 year degree & Graduate degree \\ \midrule
D1 & 1.9\% & 18.4\% & 23.3\% & 5.8\% & 25.2\% & 25.2\% \\
D2 & --    & 25.1\% & 21.4\% & 8.4\% & 21.4\% & 23.8\% \\
D3 & --    & 13.6\% & 13.6\% & 14.8\% & 19.8\% & 38.3\% \\ \bottomrule
\end{tabular}

\end{table}
\section{Ethical Statement}

Our paper studies the quality of generated images. As a measurement paper, our work provides additional information to machine learning practitioners and researchers. We hope our work will provide some transparency with regards to the efficacy of generative models in face generation. How machine learning practitioners, particularly those using generative AI, will use our results is impossible to predict or control.

Concerning user studies, our survey is IRB-approved. Furthermore, survey respondents are compensated for their time at a \$14/hr rate. Hence, we consider the human participation in our studies to be appropriate.

\end{document}